\documentclass[11pt]{article}
\usepackage[final]{acl}
\usepackage{times}
\usepackage{latexsym}
\usepackage[T1]{fontenc}
\usepackage[utf8]{inputenc}
\usepackage{microtype}
\usepackage{booktabs}
\usepackage{graphicx}
\usepackage{amsmath}
\usepackage{amssymb}
\usepackage{amsthm}
\usepackage{bm}
\usepackage{multirow}
\usepackage{rotating}
\usepackage{algorithm}
\usepackage{algpseudocode}
\usepackage[skip=4pt]{caption}
\usepackage[table,dvipsnames]{xcolor}
\usepackage[section]{placeins}
\usepackage{cuted}
\usepackage{colortbl}
\usepackage[most]{tcolorbox}
\usepackage{fancyvrb}
\usepackage{fvextra}
\usepackage{tikz}
\usetikzlibrary{positioning,arrows.meta,fit,backgrounds,shapes.geometric}
\renewcommand{\arraystretch}{0.9}
\usepackage[table, x11names]{xcolor}
\usepackage{booktabs}
\usepackage{tabularx} 
\usepackage{array}    

\usepackage{pifont}
\definecolor{ourblue}{RGB}{30,80,140}
\definecolor{ourbg}{RGB}{232,240,250}
\definecolor{groupbg}{RGB}{245,245,247}
\definecolor{fsbg}{RGB}{255,243,229}
\definecolor{edrbg}{RGB}{233,243,225}
\definecolor{gmemHeader}{RGB}{30,80,140}
\definecolor{gmemBar}{RGB}{215,228,245}
\definecolor{gmemBg}{RGB}{250,251,253}
\definecolor{gmemBorder}{RGB}{180,200,220}
\definecolor{caseText}{RGB}{30,80,140}
\definecolor{gmemLogBg}{RGB}{252,246,246}
\definecolor{gmemLogText}{RGB}{160,40,40}
\usepackage{xcolor}
\usepackage{colortbl}
\usepackage{booktabs}
\definecolor{ablHeaderBg}{RGB}{220,230,242}
\definecolor{ablGroupBg}{RGB}{242,242,242}
\definecolor{keptC}{RGB}{46,125,50}
\definecolor{rmvdC}{RGB}{198,40,40}

\newtcolorbox{gmemorycasebox}[1]{
  enhanced, breakable,
  colback=gmemBg, colframe=gmemBorder, boxrule=0.4pt,
  arc=1.2pt, left=6pt, right=6pt, top=6pt, bottom=6pt,
  fonttitle=\bfseries\small,
  coltitle=white, colbacktitle=gmemHeader,
  title={#1}, attach boxed title to top left={xshift=4pt, yshift=-3pt},
  boxed title style={size=small, colback=gmemHeader, colframe=gmemHeader, arc=1pt}
}
\newtcolorbox{gmemorycaseboxfixed}[1]{
  enhanced,
  colback=gmemBg, colframe=gmemBorder, boxrule=0.4pt,
  arc=1.2pt, left=6pt, right=6pt, top=6pt, bottom=6pt,
  fonttitle=\bfseries\small,
  coltitle=white, colbacktitle=gmemHeader,
  title={#1}, attach boxed title to top left={xshift=4pt, yshift=-3pt},
  boxed title style={size=small, colback=gmemHeader, colframe=gmemHeader, arc=1pt}
}

\newenvironment{gmemcontent}{\par\noindent\ignorespaces}{\par}
\DefineVerbatimEnvironment{redlog}{Verbatim}{
  fontsize=\scriptsize, breaklines=true, breakanywhere=true,
  frame=single, framesep=4pt, rulecolor=\color{gmemBorder},
  fillcolor=\color{gmemLogBg}, formatcom=\color{gmemLogText}}
\DefineVerbatimEnvironment{blacklog}{Verbatim}{
  fontsize=\scriptsize, breaklines=true, breakanywhere=true}
\newtheorem{definition}{Definition}

\title{VeriTrace: Evolving Mental Models for Deep Research Agents}

\author{
  Haolang Zhao$^{1,*}$ \quad Yunbo Long$^{1,*}$ \quad Lukas Beckenbauer$^{1,2}$ \quad Alexandra Brintrup$^{1,3\dagger}$ \\
  $^{1}$Department of Engineering, University of Cambridge \\
  $^{2}$TUM School of Management, Technical University of Munich \\
  $^{3}$The Alan Turing Institute\\
  \texttt{\{hz496, yl892, ab702\}@cam.ac.uk} \quad \texttt{lukas.beckenbauer@tum.de} \\
  \vspace{0.3em}\\
  {\small $^{*}$Equal contribution. \quad $^{\dagger}$Corresponding author.}
}

\begin{document}
\maketitle

\begin{abstract}
Deep research agents face vast, interdependent, and pervasively uncertain information. Existing systems explore what evolving intermediate representations should look like, but leave their evolution to the LLM's implicit reasoning. Without explicit regulation, the intermediate layer is easily contaminated by mixed-quality information, and errors propagate along its dependencies, so model scale often ends up substituting for absent regulation. We argue that an agent's mental model should instead evolve through explicit feedback that continuously aligns task understanding with reality, and identify three regulatory loops: \emph{interpretive update}, \emph{deviation feedback}, and \emph{schema revision}. We realise this in \textbf{VeriTrace}, a cognitive-graph framework that explicitly implements the three loops. Using matched Qwen3.5-27B backbones, VeriTrace improves over the strongest matched baseline by an average of 4.82~pp on DeepResearch Bench (DRB) Insight (1.83~pp Overall) and by 5.9~pp Overall win rate on DeepConsult. With Config-DeepSeek, it achieves the strongest reproducible open-source result on DRB. The code is available at \url{https://github.com/WolfGaga/VeriTrace}
\end{abstract}

\section{Introduction}

Large language model (LLM) agents are increasingly used for long-horizon information exploration, such as synthesising multi-source evidence for open-ended research questions \citep{du2025deepresearch} and navigating scientific literature in autonomous research pipelines \citep{lu2026aiscientist}. Such tasks confront the agent with information that is both voluminous and structurally interdependent. Findings must be compacted for the agent to reason over, and dependencies among them must be preserved so that one concept's findings can inform another's inquiry. These pressures push deep research systems beyond the LLM context window toward a persistent, structured intermediate layer.

\label{sec:intro}

\begin{figure}[!t]
    \centering
    \includegraphics[width=0.49
    \textwidth]{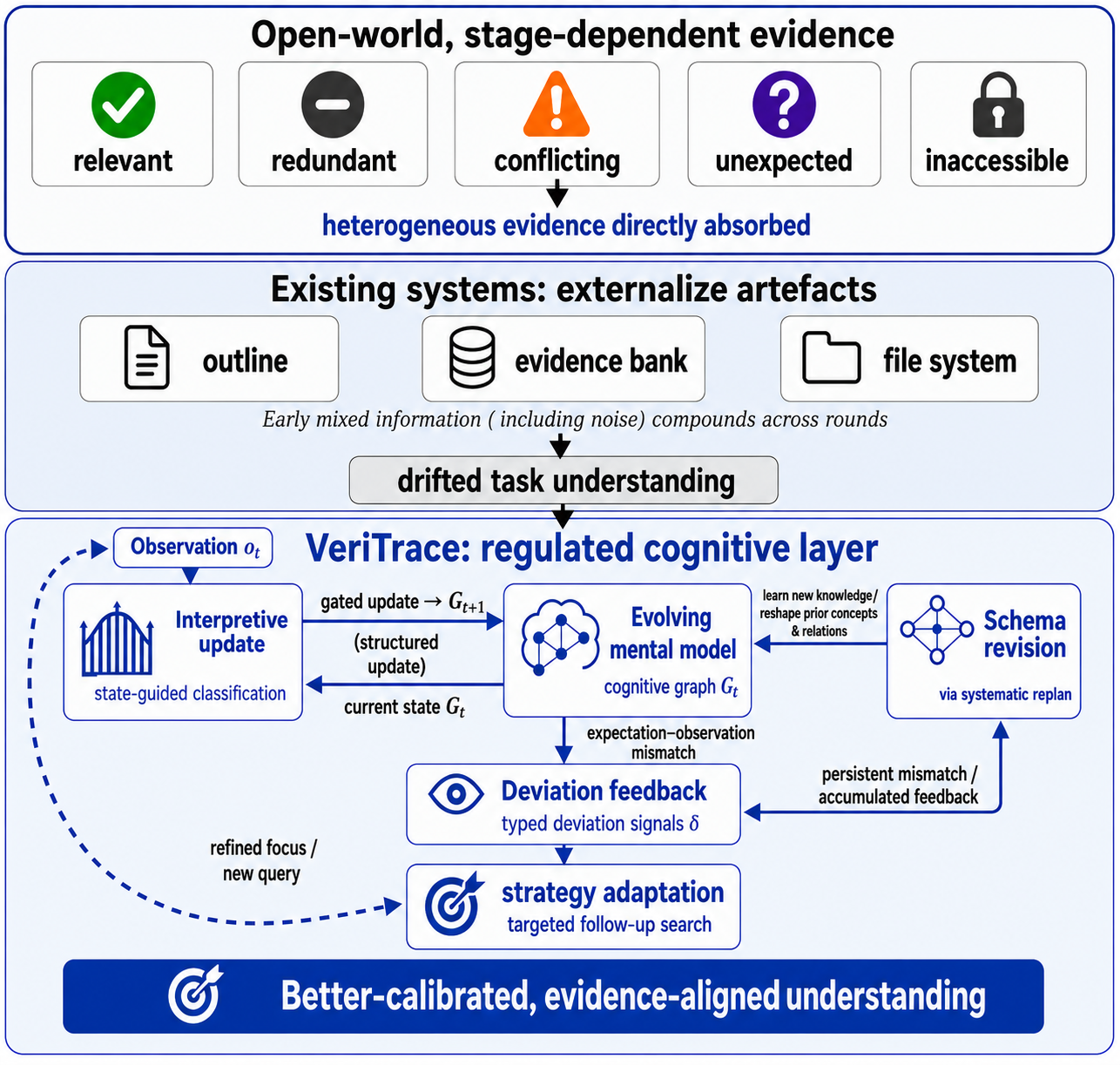}
    \caption{Regulated cognitive layer for DeepResearch.}
    \label{fig:architecture}
\end{figure}

Recent deep research systems therefore externalise intermediate research artefacts across search rounds. WebWeaver \citep{li2025webweaver} co-evolves a dynamic outline with an evidence bank. OmniThink \citep{xi2025omnithink} expands an information tree and distils a conceptual pool. FS-Researcher \citep{zhu2026fsresearcher} maintains todos and an acceptance checklist. These systems advance the question of what form an evolving intermediate layer should take, so that the agent can absorb as much knowledge as possible and organise future inquiry from it. But deep research is inherently uncertain and structurally interdependent. Without proper regulation, the intermediate layer is easily contaminated by mishandled, mixed-quality information, and errors propagate along its dependencies to distort subsequent inquiry. We therefore address a different question. What regulatory loops can enable this intermediate layer to adapt to such messiness and keep the system's task understanding evolving in a healthy direction consistent with reality?

We treat the intermediate layer as the agent's mental model of the task. This mental model is neither the transient working memory of any single decision nor a stable long-term memory that does not fully participate in decisions; it is a dynamic representation of the agent's current learning status, evidence insights, and what remains uncertain. We identify three explicit feedback loops central to this regulation (\autoref{fig:architecture}).

The first is \textbf{interpretive update}. Each new finding is read against the model already in place. As rounds accumulate, the model develops a sharper sense of what each concept already covers and still needs. An incoming finding is no longer just absorbed but classified as confirming, contradicting, gap-filling, or as a surprise outside the current framing. The state actively shapes what counts as a usable observation rather than passively accumulating them all. Without interpretive update, the system is distracted by mixed and noisy findings, loses track of its current research state and cannot target the dimensions whose evidence would most improve task understanding. Our ablation in \S\ref{sec:ablation} confirms this. Removing this loop causes the system to terminate when its store appears full rather than when concepts are actually resolved, and to dispatch fewer follow-up searches even on under-resolved concepts. 

The second is \textbf{deviation feedback}. The planner specifies what each concept expects from its next search. Each search outcome is then scored against these expectations, producing a signal that measures how far the actual evidence falls from what the planner assumed obtainable. The planner adjusts its search strategy according to this signal. Without deviation feedback, failure degenerates into undifferentiated retry rather than targeted correction. Our ablation in \S\ref{sec:ablation} confirms this. Removing this loop inflates search volume without proportional quality gain. 

The third is \textbf{schema revision}. When accumulated feedback shows the current framing itself diverges substantially from reality, the agent systematically revises its understanding of each concept and the relations among them, which in turn reshapes its plan. Otherwise the agent stays confined to its initial framing and cannot calibrate against evidence. Our ablation in \S\ref{sec:ablation} confirms this. Removing this loop noticeably degrades performance on harder, error-prone queries.

These loops are designed in response to the specific characteristics of deep research, not chosen arbitrarily. A system either implements them explicitly or delegates them to implicit LLM reasoning. The latter is a core reason deep research appears scale-hungry: model scale ends up substituting for missing regulatory loops, which is why FS-Researcher and EnterpriseDR drop substantially once the backbone is held fixed (\autoref{tab:main}).

The separation of these loops also reflects cognitive-science accounts of self-regulation. Metacognition motivates interpretive update \citep{flavell1979metacognition,nelson1990metamemory}. Predictive processing motivates deviation feedback \citep{clark2013whatever,hohwy2013predictive}. Assimilation and accommodation motivate schema revision \citep{piaget1952origins}. We treat these as design vocabulary, not cognitive-equivalence claims.

In VeriTrace, we instantiate the agent's mental model as a cognitive graph (\S\ref{sec:method}). Nodes are concepts under inquiry, edges are inquiry relations among them, and each node stores acceptance criteria, accumulated findings, and quality history. A cognitive graph manager interprets new observations against the current graph state and produces structured updates, realising interpretive update. A CR-AAP-inspired quality assessor \citep{blakeslee2004craap} measures the gap between expected and observed evidence and routes the planner to five search strategies, realising deviation feedback. When accumulated feedback shows the framing is wrong, five structural operations revise the agent's frame of understanding, and subsequent search is reorganised to test and fill the revised structure, realising schema revision.

We evaluate VeriTrace on DeepResearch Bench and DeepConsult. Under a controlled comparison with matched Qwen3.5-27B backbones, VeriTrace outperforms the strongest matched baseline by an average of 4.82 percentage points (pp) on the Insight dimension of DeepResearch Bench (1.83~pp Overall) and by 5.9~pp Overall win rate on DeepConsult. With Config-DeepSeek, VeriTrace further achieves the strongest reproducible open-source result on DeepResearch Bench (Overall 55.77, Insight 59.56).

Our contributions are threefold. \textbf{(1)} We identify three feedback loops as central regulatory mechanisms for an agent's mental model under pervasive uncertainty. \textbf{(2)} We present VeriTrace, a cognitive-graph implementation organised around concepts rather than sub-tasks, with evidence preserved under structural revision. \textbf{(3)} Comprehensive experimental studies show that explicit regulatory loops improve deep research performance beyond backbone scale alone.

\section{Related Work}
\label{sec:related}
 
\textbf{Knowledge structures in deep research systems.}
Deep research agents require intermediate artefacts that persist across search rounds and guide later search, filtering, and synthesis decisions. Existing systems differ mainly in how they materialise these artefacts. Agents like WebThinker \citep{li2025webthinker} rely on the LLM context window for implicit carry-over, while FS-Researcher \citep{zhu2026fsresearcher} drives search from todos and an acceptance checklist over a file-system store. More structured systems organise artefacts hierarchically, including WebWeaver's co-evolving outline and evidence bank \citep{li2025webweaver}, OmniThink's information tree and conceptual pool \citep{xi2025omnithink}, and Mind2Report's knowledge-enriched chapter tree \citep{cheng2026mind2report}. Our focus is not which storage form best preserves intermediate artefacts, but what regulatory conditions allow the intermediate layer to improve its task understanding over long research trajectories.
 
\textbf{Cognitive-science framing.}
We use cognitive science as a design vocabulary rather than a claim of cognitive equivalence. \emph{Metacognition} \citep{flavell1979metacognition} and the monitoring--control view of metamemory \citep{nelson1990metamemory} cast learning as monitoring what one knows and adjusting inquiry accordingly, motivating interpretive update. \emph{Predictive processing} \citep{clark2013whatever,hohwy2013predictive} frames action as driven by expectation--observation mismatch, motivating deviation feedback. Piaget's contrast between \emph{assimilation} into existing structure and \emph{accommodation} of the structure itself \citep{piaget1952origins} motivates schema revision.
 
\textbf{Replanning versus reframing}. A related line revises plans after detecting problems. EnterpriseDR \citep{prabhakar2025enterprise} reflects on gaps at the knowledge level to drive dynamic replanning, accumulating knowledge accordingly. Schema revision instead restructures the concept labels, their relations, and the inquiry goals that together define how evidence insights are organised and where search goes next, and then reorganises search to fill and test the revised frame. Thus, a framing error is corrected, rather than diluted by accumulating knowledge within the wrong frame (\S\ref{sec:ablation}).

\section{Method}
\label{sec:method}

\begin{figure*}[t]
    \centering
    \includegraphics[width=1.05\textwidth]{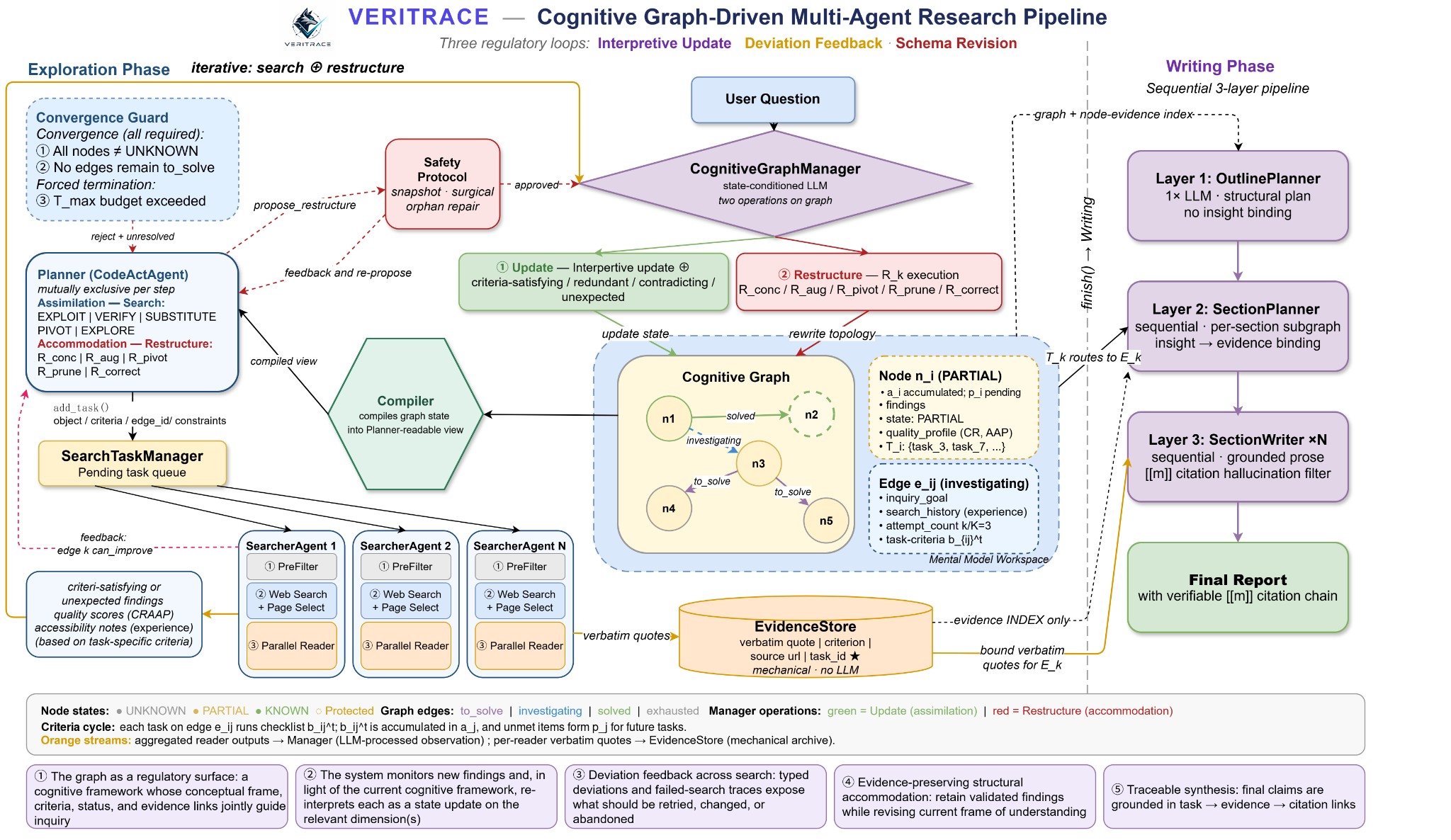}
    \caption{VeriTrace architecture. The cognitive graph coordinates regulated exploration and traceable synthesis.}
     \label{fig:method_architecture}
\end{figure*}

\subsection{Overview}
\label{sec:overview}

VeriTrace centres the research loop on a cognitive graph (Figure~\ref{fig:method_architecture}). A \textbf{planner} reads a compiled view of the graph state (Appendix~\ref{app:cg_schema}) and issues structured actions, while parallel \textbf{searchers} gather evidence and a lightweight \textbf{reader} extracts findings with page-level quality scores. The \textbf{cognitive graph manager} then assimilates observations into the graph state (Appendix~\ref{app:update}). The loop alternates between assimilation and accommodation. Unresolved edges trigger search, deviation feedback routes follow-up strategies (\S\ref{sec:strategy}), and restructuring is invoked when accumulated feedback no longer fits the current frame of understanding (\S\ref{sec:accommodation}, \S\ref{sec:dynamics}). Once all nodes leave \emph{unknown} and all edges are resolved, the writing pipeline produces the final report from the graph and evidence store (\S\ref{sec:writing}, with Appendix~\ref{app:closure_matrix} giving a feature-level comparison against open-source baselines on the three loops).

\subsection{Cognitive Graph}
\label{sec:cg}

\begin{definition}[Cognitive Graph]
\label{def:cg}
A cognitive graph at time $t$ is a tuple $G_t = (S_t, C_t)$. The \textbf{structural state} $S_t = (N_t, E_t, N_{\textup{user}})$ consists of a node set $N_t$, an edge set $E_t \subseteq N_t \times N_t$, and a protected subset $N_{\textup{user}} \subseteq N_t$ (nodes for dimensions explicitly specified in the user's question); the \textbf{content state} $C_t = (C_t^N, C_t^E)$ consists of node content $C_t^N = \{c_i^t\}_{n_i \in N_t}$ and edge content $C_t^E = \{c_{ij}^t\}_{e_{ij} \in E_t}$.
\end{definition}

\begin{definition}[Concept Node Content]
\label{def:node}
Each node $n_i \in N_t$ carries $c_i^t = (a_i^t, f_i^t, s_i^t, \bm{q}_i^t, \mathcal{T}_i^t)$: acceptance criteria $a_i^t \in \mathcal{L}$ that have not yet been satisfied by current findings $f_i^t$, findings $f_i^t$ (a set of atomic findings, monotonically accumulated on the regular update path), cognitive state $s_i^t \in \{\textsc{unknown}, \textsc{partial}, \textsc{known}\}$, quality profile $\bm{q}_i^t = (\overline{\textup{cr}}_i^t, \overline{\textup{aap}}_i^t) \in [1,5]^2$, and associated task IDs $\mathcal{T}_i^t \subseteq \mathbb{N}$.
Here $\mathcal{L}$ is the space of natural language descriptions. The cognitive state is determined by:

{\small
\begin{equation}
    s_i^t = \sigma(a_i^t, f_i^t) = \begin{cases}
        \textsc{known}   & a_i^{\textup{core},t} \subseteq \textup{sat}(f_i^t) \\
        \textsc{partial} & \begin{gathered}
            a_i^{\textup{core},t} \not\subseteq \textup{sat}(f_i^t) \\
            {}\wedge\ f_i^t \neq \emptyset
        \end{gathered} \\
        \textsc{unknown} & f_i^t = \emptyset
    \end{cases}
\end{equation}
}%
where $a_i^{\textup{core},t}$ is the core criteria subset at $t$ and $\textup{sat}(f_i^t)$ is the set of criteria satisfied by current findings. $\sigma$ is realised as a structured LLM judgement (Appendix~\ref{app:update}).
\end{definition}

\begin{definition}[Edge Content]
\label{def:edge}
Each edge $e_{ij} \in E_t$ carries
$c_{ij}^t = (g_{ij}, b_{ij}^t, h_{ij}^t, k_{ij}^t, u_{ij}^t)$:
inquiry goal $g_{ij} \in \mathcal{L}$ as the strategic question this edge investigates, task-specific acceptance criteria $b_{ij}^t \in \mathcal{L}$ that break $g_{ij}$ into checkable items for the most recent task on this edge, search history $h_{ij}^t \in \mathcal{L}^*$, attempt count
$k_{ij}^t \in \{0, \ldots, K\}$ ($K{=}3$), and status $u_{ij}^t$
taking one of \emph{to-solve}, \emph{investigating}, \emph{solved}, or
\emph{exhausted}.
\end{definition}

\subsection{Content Evolution (Assimilation)}
\label{sec:assimilation}

At each search step, the planner selects an unresolved edge $e_{ij}$ from $G_t$ and dispatches a search task (full task schema in Appendix~\ref{app:prompts}, Planner block). For each task, the planner instantiates $b_{ij}^t$ as concrete sub-criteria of $g_{ij}$, drawing on $a_j$ to target what node $n_j$ still owes. These criteria serve as the \emph{prior expectation} that the searcher executes and the reader extracts and scores findings against. A searcher executes the task with access to the task specification and local context (the target node and its upstream neighbours, not the full graph); for each retrieved page, a reader extracts findings and computes per-page CR-AAP quality scores. The combined output forms the observation $o_t = (\hat{f}_t,\, \hat{\bm{q}}_t)$, where $\hat{f}_t$ are extracted findings and $\hat{\bm{q}}_t$ are quality scores.
The cognitive graph manager then interprets $o_t$ through the lens of $G_t$, routing each finding to $n_j$ by default and to $n_k$ ($k \neq j$) when it aligns with $a_k$. This prior-guided interpretation is the core of assimilation. Each relevant finding is classified against $b_{ij}^t$ and $f_j^t$ as \emph{criterion-satisfying}, \emph{redundant}, \emph{contradictory}, or \emph{unexpected}, where unexpected requires both relevance to the user's original question on $n_j$'s dimension and absence from every entry of $b_{ij}^t$. At task closure, entries of $b_{ij}^t$ left unsatisfied by $o_t$ are retained in $a_j^{t+1}$, so $n_j$ carries them across edges as residual criteria shaping its next task.

The resulting update set $\mathcal{U}_t = \{(j, \Delta c_j^t)\}$ thus contains $n_j$ by default and \emph{may} span additional nodes:

{\footnotesize
\begin{equation}
\label{eq:content_update}
\forall (j,\Delta c_j^t)\in\mathcal U_t:\quad
\left\{
\begin{aligned}
f_j^{t+1} &= f_j^t \oplus_G \Delta f_j^t,\\
\bm q_j^{t+1} &= \operatorname{agg}(\bm q_j^t,\Delta\bm q_j^t),\\
s_j^{t+1} &= \sigma(a_j^{t+1},f_j^{t+1}).
\end{aligned}
\right.
\end{equation}
}
Nodes not referenced in $\mathcal{U}_t$ retain their state. Equation~\ref{eq:content_update} realises \textbf{interpretive update}: $\oplus_G$ folds each classified finding into the node's two-level finding structure (\emph{item\_findings} or \emph{cross\_item\_findings}) rather than a flat list, then recomputes $s_j^{t+1} = \sigma(a_j^{t+1}, f_j^{t+1})$ so the system tracks how far each concept is resolved. Redundant findings are not re-added; contradictions are recorded in a per-criterion contradictions list preserving prior evidence; unexpected discoveries are routed to a dedicated channel. The latter two are surfaced as candidate triggers for \textbf{schema revision} (Appendix~\ref{app:update}).

When a search task for $e_{ij}$ contributes findings to node $n_k$ ($k \neq j$), the task identifier is registered in $n_k$'s associated task set: $\mathcal{T}_k^{t+1} = \mathcal{T}_k^t \cup \{\textup{task}_t\}$. This cross-node linkage is essential for the writing pipeline (\S\ref{sec:writing}).

\subsection{Deviation Detection and Strategy Selection}
\label{sec:strategy}

\begin{definition}[Deviation Signal]
\label{def:deviation}
Upon completing a search targeting node $n_j$, the system computes a deviation signal in $\mathcal{D} \triangleq [1,5]^2 \times \{0,1\} \times [1,4]$:

{\small
\begin{equation}
    \delta_j^t = \big(\,\overline{\textup{cr}}_j^t,\;\; \overline{\textup{aap}}_j^t,\;\; \phi_j^t,\;\; \psi_j^t\,\big) \;\in\; \mathcal{D}
\end{equation}
}%
with content relevance $\overline{\textup{cr}} \in [1,5]$ (evidence--need alignment), source credibility $\overline{\textup{aap}} \in [1,5]$ (\emph{independent of} relevance), accessibility barrier $\phi \in \{0,1\}$, and unexpected-finding strength $\psi \in [1,4]$. The orthogonality of $\overline{\textup{cr}}$ and $\overline{\textup{aap}}$ enables differential diagnosis (Appendix~\ref{app:craap}).
\end{definition}

The planner selects a search strategy through a mapping $\pi: \mathcal{D} \to \mathcal{A}_\textup{search}$ that partitions the deviation space into five regions. We define predicates \textsc{Accessible}$(\delta) \triangleq (\phi = 0)$, \textsc{Rel}(evant)$(\delta) \triangleq (\overline{\textup{cr}} \geq \tau_h)$, \textsc{Cred}(ible)$(\delta) \triangleq (\overline{\textup{aap}} \geq \tau_h)$, and \textsc{Unex}(pected)$(\delta) \triangleq (\psi \geq \tau_\psi)$. The mapping is (evaluated in priority order):

{\footnotesize
\begin{equation}
    \pi(\delta) = \begin{cases}
        \textsc{substitute} & \neg\,\textsc{Accessible} \\
        \textsc{exploit}    & \textsc{Rel} \wedge \textsc{Cred} \\
        \textsc{verify}     & \textsc{Rel} \wedge \neg\,\textsc{Cred} \\
        \textsc{pivot}      & \neg\,\textsc{Rel} \wedge \textsc{Unex} \\
        \textsc{explore}    & \text{otherwise}
    \end{cases}
    \label{eq:strategy}
\end{equation}
}%
\emph{exploit} refines within the current source family; \emph{verify} cross-checks against authoritative sources; \emph{substitute} switches source type when blocked; \emph{pivot} routes an unexpected finding to a new edge; \emph{explore} broadens keywords or language. Threshold values and execution details are in Appendix~\ref{app:strategies}.

\subsection{Framework Evolution (Accommodation)}
\label{sec:accommodation}
When repeated strategy adaptations on an edge leave $\delta_j^t$ in the same deviation region, or when contradictions accumulate across the graph, the planner concludes that the current framing itself is misaligned rather than merely under-informed, and proposes a restructuring operation $R_k \in \mathcal{A}_\textup{struct}$ that revises the frame of understanding (concept labels, relations, inquiry goals, and criteria), thereby influencing the planner's future inquiry plans while preserving evidence under invariant \textbf{I1} (\S\ref{sec:accommodation}):

{\footnotesize
\begin{equation}
    S_{t+1} = R_k(S_t, C_t), \;\; \mathcal{A}_\textup{struct} = \left\{\!\begin{array}{l} R_\textup{conc}, R_\textup{aug}, R_\textup{pivot},\\ R_\textup{prune}, R_\textup{correct}\end{array}\!\right\}
\end{equation}
}%
The equations below describe the semantic effect of each $R_k$ on $S_t$;
the concrete edits, including free symbols such as $E_\textup{old}$, $E_\textup{reorg}$, $N_\textup{abs}$, and $N_\textup{item}$, are produced at execution time by a separate graph-manager component under a multi-phase safety protocol (Appendix~\ref{app:safety}).

\textbf{Local effects} on nodes/edges:

{\footnotesize
\begin{align}
    R_\textup{conc}(n_i) &: N_{t+1} \!=\! N_t \cup \{n_{i_1}, \ldots, n_{i_m}\},\nonumber\\
                         &\quad E_{t+1} \!=\! E_t \cup \{(n_i, n_{i_k})\}_{k=1}^m \label{eq:conc} \\
    R_\textup{aug} &: N_{t+1} \!=\! N_t \cup \{n_\textup{new}\},\; E_{t+1} \!=\! E_t \cup \{e_\textup{new}\} \label{eq:aug} \\
    R_\textup{pivot} &: N_{t+1} \!=\! (N_t \setminus N_\textup{abs}) \cup N_\textup{item},\nonumber\\
                     &\quad E_{t+1} \!=\! (E_t \setminus E_\textup{old}) \cup E_\textup{reorg} \label{eq:pivot} \\
    R_\textup{prune}(n_i) &: N_{t+1} \!=\! N_t \!\setminus\! \{n_i\},\; E_{t+1} \!=\! E_t \!\setminus\! \{e \!\ni\! n_i\},\nonumber\\
                          &\quad n_i \!\notin\! N_\textup{user},\; f_i^t \!=\! \emptyset \label{eq:prune}
\end{align}
}%
\textbf{Cascading effect.} When evidence at $n_i$ contradicts a premise
on which downstream nodes were planned, $R_\textup{correct}$ targets the
affected subgraph $\downarrow\!n_i \triangleq \{n_j \in N_t \mid n_i \leadsto n_j\}$.
Inquiry plans ($g_{kj}, a_j$) on $\downarrow\!n_i$ are revised subject to \textbf{I1} (\S\ref{sec:accommodation}):

{\small
\begin{equation}
    R_\textup{correct}(n_i) : S_t\big|_{\downarrow n_i} \;\longmapsto\; S_t'
    \label{eq:correct}
\end{equation}
}%
\textbf{Structural invariants.} Let $N_\textup{user}$ denote the nodes corresponding to dimensions explicitly named in the user query. All $R_k \in \mathcal{A}_\textup{struct}$ satisfy:

{\footnotesize
\begin{align}
    \textbf{I1} &\;\text{(Immutable Past):}\; \forall n_j\!: f_j^t \!\neq\! \emptyset \Rightarrow f_j^{t+1} \!\supseteq\! f_j^t \label{eq:immutable} \\
    \textbf{I2} &\;\text{(Dim.\ Protection):}\; N_\textup{user} \subseteq N_{t+1} \label{eq:protect}
\end{align}
}%
\textbf{I1} is enforced by append-only updates to $f_j^t$, with one documented exception for noise-item cleanup (Appendix~\ref{app:safety}). \textbf{I2}
is enforced as a veto condition during execution: any proposal violating
$N_\textup{user}$ membership is rejected (Appendix~\ref{app:safety}).

\subsection{System Dynamics}
\label{sec:dynamics}

At each time step, the planner inspects $G_t$ and decides between two mutually exclusive actions: dispatching a search task (assimilation) or proposing a restructuring operation (accommodation). The two never co-occur within a single step:

{\footnotesize

\begin{equation}
\boxed{
    G_{t+1} \!=\! \begin{cases}
        \bigl(S_t,\; \Gamma(G_t, o_t)\bigr) & \!\!\text{search step} \\[4pt]
        \bigl(R_k(S_t, C_t),\; C_t\bigr) & \!\!\text{restructuring step}
    \end{cases}
}
\label{eq:dynamics}
\end{equation}
}%
In the common case, the planner dispatches search tasks and the system assimilates observations into $C_t$. Deviation signals $\delta_j^t$ accumulate over successive search steps; the planner uses these to select search strategies (\S\ref{sec:strategy}) and attempts to resolve deviations through strategy adaptation. Only when accumulated signals indicate that the current frame itself is fundamentally wrong does the planner forgo search and instead propose a restructuring operation, modifying $S_t$ while leaving $C_t$ unchanged. After restructuring, the system resumes assimilation on the updated topology.

\textbf{Termination.}
The search loop terminates at step $T$ when:

{\footnotesize
\begin{align}
    &\forall\, n_i \in N_T:\; s_i^T \neq \textsc{unknown} \nonumber \\
    \wedge\;\; &\nexists\; e_{ij} \in E_T:\; u_{ij}^T = \textsc{to-solve}\nonumber
\end{align}
}
Nodes leave \emph{unknown} only when the planner's scheduling reaches their edges; topologically distant nodes thus remain \emph{unknown} until selected. A soft budget $T_\textup{max}$ enforces termination if convergence is not reached (Appendix~\ref{app:termination}).

\subsection{Writing Pipeline}
\label{sec:writing}

Upon termination, the cognitive graph $G_T$ and an evidence store $\mathcal{E}$ jointly feed a three-layer writing pipeline. The evidence store accumulates structured records extracted by readers throughout the search loop:

{\small
\begin{equation}
    \mathcal{E} = \bigl\{\, (m,\, v_m,\, w_m,\, \ell_m,\, \textup{src}_m,\, \textup{task}_m) \,\bigr\}_{m=1}^{M}
\end{equation}
}%
where $v_m \in \mathcal{L}$ is a verbatim quote from the source page, $w_m \in \mathcal{L}$ is the reader's finding summary, $\ell_m \in \mathcal{L}$ is the acceptance criterion this finding addresses, $\textup{src}_m$ is the source URL, and $\textup{task}_m$ is the originating search task identifier. A single source page may produce multiple records. Records enter $\mathcal{E}$ without further LLM processing; this mechanical side-channel ensures citation traceability.

 Each node $n_k$'s evidence subset $\mathcal{E}_k \subseteq \mathcal{E}$ collects records whose $\textup{task}_m \in \mathcal{T}_k$. The pipeline narrows information and decision authority in lockstep (Appendix~\ref{app:writing}):
\textbf{Outline planner}: reads $G_T$ with findings $f_k^T$; outputs sections, each mapped to a node subset $V_k \subseteq N_T$.
\textbf{Section planner} (sequential): reads $V_k$'s findings, the index $\{(m, \ell_m) : m \in \bigcup_{n_k \in V_k}\mathcal{E}_k\}$ (each $\ell_m$ already names an acceptance criterion of the section), and prior sections' plans; produces \emph{insights} bound to 2--5 evidence IDs. \textbf{Section writer} (sequential): receives insights with bound records from $\bigcup_{n_k \in V_k}\mathcal{E}_k$ unlocked; renders prose with $[\![m]\!]$ citations.

\section{Experimental Setup}
\label{sec:setup}

\subsection{Benchmarks}

We evaluate on \textbf{DeepResearch Bench} \citep{du2025deepresearch}, comprising 100 complex research queries in Chinese and English. It employs two evaluation frameworks: (1)~\textbf{RACE}, which assesses report quality via four LLM-judged dimensions (Comprehensiveness, Insight, Instruction Following, and Readability), scored relative to expert reference reports (0.5 = reference parity); and (2)~\textbf{FACT}, which measures citation reliability through Effective Citations and Citation Accuracy. We additionally evaluate on \textbf{DeepConsult} \citep{DeepConsult}, a 102-query pairwise benchmark judged by \texttt{gemini-2.5-pro}, to test cross-benchmark transfer; results are reported in \autoref{tab:deepconsult_main}.

\subsection{Baselines and Model Configuration}

We compare against two categories of systems. \textbf{Proprietary systems} include Claude-DeepResearch, OpenAI-DeepResearch, Gemini-2.5-Pro-DeepResearch, Doubao-DeepResearch, and Kimi-Researcher, with results cited from official leaderboards as their implementations are not publicly available. \textbf{Open systems with disclosed methodology} include LangChain Open Deep Research, EnterpriseDR, WebWeaver, RhinoInsight, and FS-Researcher, each evaluated under its best reported backbone. For the controlled comparison (\autoref{tab:main}), we further rerun the three systems with both released code and a published paper (WebWeaver, EnterpriseDR, and FS-Researcher) under our shared Qwen3.5-27B configuration to isolate architectural contributions from model capability. The per-role model configurations used by VeriTrace are shown in \autoref{tab:configs}.

\begin{table}[t]
\centering
\fontsize{7.5}{8.5}\selectfont
\setlength{\tabcolsep}{3pt}
\renewcommand{\arraystretch}{0.95}
\begin{tabular*}{\columnwidth}{@{\extracolsep{\fill}} llll @{}}
\toprule
\rowcolor{groupbg} \textbf{Config} & \textbf{P/S} & \textbf{Reader} & \textbf{Writer} \\
\midrule
Config-3B        & Q3.5-35B-A3B & Q3.5-35B-A3B   & Q3.5-35B-A3B \\
Config-27B       & Q3.5-27B     & Q3.5-35B-A3B   & Q3.5-27B \\
Config-Plus      & Q3.5-Plus    & Q3.5-122B-A10B & Q3.6-Plus \\
Config-DeepSeek  & DSv4-Pro    & DSv4-Flash     & DSv4-Pro \\
\bottomrule
\end{tabular*}
\caption{Q = Qwen, DSv4 = DeepSeek-V4. P/S = Planner/Searcher. The Cognitive Graph Manager uses the P/S model in Qwen \citep{yang2025qwen3} configurations and the Reader's model in Config-DeepSeek \citep{deepseekai2026deepseekv4}. PreFilter details are in Appendix~\ref{app:prefilter}.}
\label{tab:configs}
\end{table}

\section{Results}
\label{sec:results}

\subsection{Main Results}

\autoref{tab:leaderboard} reports DeepResearch Bench performance. Within the Qwen family, VeriTrace remains close across three configurations, with Overall scores ranging from 51.96 to 53.50. Switching to a different open-source backbone family with Config-DeepSeek lifts Overall to 55.77, giving the strongest reproducible open-source result in the table. On the \emph{Insight} dimension, Config-DeepSeek (59.56) and Config-Plus (57.08) achieve the best and second-best scores across all listed systems.

\begin{table*}[t] 
\centering
\rowcolors{2}{white}{white} 
\fontsize{9}{11}\selectfont
\setlength{\tabcolsep}{3pt} 
\renewcommand{\arraystretch}{0.95}

\begin{tabularx}{\textwidth}{l *{7}{>{\centering\arraybackslash}X}}
\toprule
\rowcolor{groupbg}
\textbf{Method} & \textbf{Comp.} & \textbf{Insight} & \textbf{Instr.} & \textbf{Read.} & \textbf{Overall} & \textbf{Eff.c.} & \textbf{C.acc.} \\
\midrule
Claude-DeepResearch \citep{anthropic2025research} & 45.34 & 42.79 & 47.58 & 44.66 & 45.00 & -- & -- \\
OpenAI-DR \citep{openai2025deepresearch} & 46.46 & 43.73 & 49.39 & 47.22 & 46.45 & 39.79 & 75.01 \\
Gemini-2.5-Pro-DR \citep{google2025geminidr} & 49.51 & 49.45 & 50.12 & 50.00 & 49.71 & 165.34 & \textbf{78.30} \\
Doubao-DeepResearch \citep{doubao2025deepresearch} & 44.84 & 40.56 & 47.95 & 44.69 & 44.34 & 52.62 & 52.86 \\
Kimi-Researcher \citep{moonshot2025kimiresearcher} & 44.96 & 41.97 & 47.14 & 45.59 & 44.64 & -- & -- \\
\midrule
LangChain-Open-Deep-Research(GPT-5) \citep{langchain2025odr} & 50.06 & 50.76 & 51.31 & 49.72 & 50.60 & 22.44 & 34.74 \\
EnterpriseDR(Gemini-2.5-Pro) \citep{prabhakar2025enterprise} & 49.70 & 51.24 & 50.52 & 50.61 & 50.62 & -- & 72.50 \\
WebWeaver (Qwen3-235B-A22B-Instruct-2507) \citep{li2025webweaver} & 51.29 & 51.00 & 49.98 & 48.89 & 50.62 & \underline{166.73} & \underline{78.25} \\
RhinoInsight(Gemini-2.5-Pro) \citep{lei2025rhinoinsight} & 50.51 & 51.45 & 51.72 & 50.00 & 50.92 & -- & -- \\
FS-Researcher (Claude-Sonnet-4.5) \citep{zhu2026fsresearcher} & \underline{54.25} & 55.85 & \textbf{52.47} & \underline{51.54} & \underline{53.94} & 139.91 & 76.17 \\
\midrule
\rowcolor{ourbg} VeriTrace (Config-3B) & 50.85 & 55.22 & 50.38 & 49.08 & 51.96 & -- & -- \\
\rowcolor{ourbg} VeriTrace (Config-27B)\textsuperscript{*} & 51.51 & 56.12 & 50.64 & 49.87 & 52.60 & 134.20 & 68.46 \\
\rowcolor{ourbg} VeriTrace (Config-Plus) & 52.84 & \underline{57.08} & 51.65 & 49.61 & 53.50 & -- & -- \\
\rowcolor{ourbg} {VeriTrace (Config-DeepSeek)} & \textbf{56.28} & \textbf{59.56} & \underline{52.24} & \textbf{51.98} & \textbf{55.77} & \textbf{235.66} & 73.28 \\
\bottomrule
\end{tabularx}
\caption{DeepResearch Bench performance. Bold and underline indicate best and second-best scores. \textsuperscript{*} Config-27B RACE scores are means across three independent full runs; FACT scores are from the first run.}
\label{tab:leaderboard}
\end{table*}

To isolate architectural contributions from model capability, \autoref{tab:main} compares systems under the shared Qwen3.5-27B backbone. VeriTrace achieves the highest average Overall score ($+1.83$~pp over WebWeaver) and leads all matched baselines on Insight, Instruction Following, and Readability, with its largest margin on Insight ($+4.82$~pp).

\begin{table*}[t]
\centering
\fontsize{9}{11}\selectfont
\setlength{\tabcolsep}{6pt}
\renewcommand{\arraystretch}{0.95}
\begin{tabularx}{\textwidth}{l *{5}{>{\centering\arraybackslash}X}}
\toprule
\rowcolor{groupbg}
\textbf{System (Qwen3.5-27B)} 
& \textbf{Comp.} & \textbf{Insight} & \textbf{Instr.} & \textbf{Read.} & \textbf{Overall} \\
\midrule
WebWeaver & \textbf{51.83$\pm$0.50} & 51.30$\pm$0.33 & 49.60$\pm$0.47 & 49.23$\pm$0.41 & 50.77$\pm$0.34 \\
FS-Researcher & 47.83 & 47.05 & 48.74 & 49.37 & 48.03 \\
Enterprise-DR & 47.82 & 50.17 & 49.30 & 48.91 & 49.16 \\
\midrule
\rowcolor{ourbg}
{VeriTrace (Config-27B)} 
& 51.51$\pm$0.65 & \textbf{56.12$\pm$0.26} & \textbf{50.64$\pm$0.15} & \textbf{49.87$\pm$0.50} & \textbf{52.60$\pm$0.31} \\
\bottomrule
\end{tabularx}
\caption{Controlled RACE comparison under the shared Qwen3.5-27B backbone. VeriTrace and WebWeaver report mean $\pm$ standard deviation across three independent full runs.}
\label{tab:main}
\end{table*}

To test whether the architectural gain transfers beyond DeepResearch Bench, we also evaluate on DeepConsult under the shared Qwen3.5-27B backbone. As shown in \autoref{tab:deepconsult_main}, VeriTrace reaches 81.1\% Overall win rate, +5.9~pp over the strongest matched baseline (WebWeaver, 75.2\%), with higher average score and NWR. Per-dimension results are reported in Appendix~\ref{app:Deepconsult}. Separately, a blind evaluation by four non-author PhD annotators on 20 DeepResearch Bench queries yielded positive net win rates for VeriTrace for all four annotators ($+30\%$ to $+56\%$). The strongest preference was observed on Insight (Appendix~\ref{app:human_eval}).



\begin{table}[t]
\centering
\fontsize{9}{11}\selectfont
\setlength{\tabcolsep}{4pt}
\renewcommand{\arraystretch}{0.95}
\begin{tabularx}{\columnwidth}{l *{3}{>{\centering\arraybackslash}X}}
\toprule
\rowcolor{groupbg}
\textbf{System} & \textbf{Win\%} & \textbf{Avg.\ Score} & \textbf{NWR} \\
\midrule
Enterprise-DR & 64.7 & 6.11 & 0.769 \\
FS-Researcher & 72.8 & 6.49 & 0.797 \\
WebWeaver     & 75.2 & 6.81 & 0.859 \\
\midrule
\rowcolor{ourbg}{VeriTrace} 
& \textbf{81.1} & \textbf{6.94} & \textbf{0.887} \\
\bottomrule
\end{tabularx}
\caption{DeepConsult Overall under Qwen3.5-27B (102 queries; judge: \texttt{gemini-2.5-pro}). Avg.\ Score is 0--10 with 5 denoting a tie; NWR is computed over non-tie outcomes.}
\label{tab:deepconsult_main}
\end{table}

\subsection{Ablation Study}
\label{sec:ablation}

We design four ablations targeting the three loops from \S\ref{sec:intro} and the topological substrate that schema revision operates on. \textbf{A1} (\emph{deviation feedback}) removes the deviation signal $\delta$ and strategy selector $\pi$, forcing the planner to decide search directions from implicit reasoning over raw findings. \textbf{A2} (\emph{schema revision}) restricts $\mathcal{A}_\textup{struct}$ to $\{R_\textup{aug}, R_\textup{prune}\}$, the subset already available in existing systems. \textbf{A3} (\emph{interpretive update}) replaces $\oplus_G$ with naive text concatenation. \textbf{A4} (\emph{topological substrate}) flattens the cognitive graph to a parallel list of dimensions, removing edges and hop distances. \textbf{A\textsubscript{full}} stacks all four. Implementation details are in Appendix~\ref{app:ablation-impl}.

\begin{table}[t]
\centering
\fontsize{7.5}{8.5}\selectfont
\setlength{\tabcolsep}{1.5pt}
\renewcommand{\arraystretch}{0.95}
\begin{tabular*}{\columnwidth}{@{\extracolsep{\fill}} lccccc @{}}
\toprule
\rowcolor{groupbg}
\textbf{System} & \textbf{Overall} & \textbf{$\Delta$} & \textbf{W/L} & \textbf{Srch.} & \textbf{Succ.} \\
\midrule
\rowcolor{ourbg}
Full VeriTrace (baseline) 
& \textbf{52.60$\pm$0.31} & --- & --- & 1.00$\times$ & 100/100 \\
$-$ Deviation feedback (A1)  
& 50.52 & $-$1.96 & 14/72 & 1.22$\times$ & 86/100 \\
$-$ Schema revision (A2)    
& 51.28 & $-$1.36 & 33/63 & 0.98$\times$ & 96/100 \\
$-$ Interpretive update (A3)    
& 50.75 & $-$1.83 & 22/77 & 0.41$\times$ & 99/99 \\
$-$ Graph topology (A4)  
& 51.39 & $-$1.20 & 33/66 & 1.14$\times$ & 99/99 \\
$-$ All four (A\textsubscript{full}) 
& 49.80 & $-$2.81 & 3/97 & 0.54$\times$ & 100/100 \\
\bottomrule
\end{tabular*}
\caption{Ablation on RACE Overall. Full: mean $\pm$ standard deviation across three runs; $\Delta$: mean per-query difference (ablation $-$ three-run Full); \textbf{W/L}: wins/losses vs.\ three-run Full; \textbf{Srch.}: search-call ratio vs.\ three-run Full; \textbf{Succ.}: queries finished within 90~min.}
\label{tab:ablation}
\end{table}

\noindent\textbf{Accommodation-sensitive subset.}
We first focus on queries with $p\geq2/3$, identified across the three full runs. These form the \emph{accommodation-sensitive subset} (\autoref{tab:ablation-accomm}). Here, A2, A3, and A\textsubscript{full} show similarly severe losses ($-4.61$, $-4.87$, and $-4.73$~pp), while A4 is lighter ($-2.98$~pp). These contrasts raise two questions: why is A3 worse than A2, and why is A4 lighter than A2?

\begin{table}[t]
\centering
\fontsize{9}{11}\selectfont
\setlength{\tabcolsep}{0.8pt}
\renewcommand{\arraystretch}{0.95}
\begin{tabular*}{\columnwidth}{@{\extracolsep{\fill}} lcccc @{}}
\toprule
\rowcolor{groupbg}
\textbf{Ablation} & \textbf{$p=0$} & \textbf{$p=1/3$} & \textbf{$p\geq2/3$} & \textbf{Wilcoxon $p_{\mathrm{W}}$} \\
\midrule
A1 & $-1.96$ (65) & $-2.39$ (12) & $-1.37$ (9) & .496 \\
A2 & $-0.92$ (72) & $-1.52$ (15) & $-4.61$ (9) & .008 \\
A3 & $-1.20$ (74) & $-2.82$ (14) & $-4.87$ (11) & .010 \\
A4 & $-1.04$ (74) & $-0.64$ (14) & $-2.98$ (11) & .014 \\
A\textsubscript{full} & $-2.14$ (74) & $-4.73$ (15) & $-4.73$ (11) & .003 \\
\bottomrule
\end{tabular*}
\caption{Propensity $p$ is the fraction of three full runs in which a query triggered $R_\textup{conc}$, $R_\textup{pivot}$, or $R_\textup{correct}$. For each propensity group, $\Delta$ is calculated by subtracting the three-run mean full-system score from the ablation score for each query and then averaging these differences over all scored queries in the group. Parentheses show the number of scored queries. The final column gives $p_{\mathrm{W}}$-values from two-sided Wilcoxon signed-rank tests of whether matched $\Delta$ values in the $p\geq2/3$ subset are centred at zero.}
\label{tab:ablation-accomm}
\end{table}

\noindent\textbf{Why A3 is worse than A2.}
A3 (no interpretive update) should not, in principle, be tied to schema revision. Yet traces show that A3 still triggers restructuring operations, and on the accommodation-sensitive subset it falls further than A2, which directly disables the $R$-operations. Without interpretive update maintaining a coherent state, the restructuring that does fire becomes a side effect rather than a correction. Interpretive update provides the cognitive basis on which $R$-operations can make informed judgments; it is the precondition for schema revision to act correctively at all. The same ordering holds across all three propensity levels. A3 is consistently worse than A2, and both losses deepen as $p$ rises.

\noindent\textbf{Why A4 is lighter than A2.}
A4 (flattened to a list) is in fact lighter on the accommodation-sensitive subset than A2 (graph with restricted R-operations). This is not evidence that lists outperform graphs: full VeriTrace still outperforms A4 overall (\autoref{tab:ablation}). The reason is specific: dependency-bearing topology better captures conceptual dependencies, but under long-horizon uncertainty small models misjudge such structures, and a single misjudgment cascades along dependencies, making the graph \emph{less} fault-tolerant than a flat list. Paired with an adequate repair mechanism, this fault-tolerance penalty is removed and the graph's inherent advantage is restored. This sharpens what schema revision is for: the precondition that lets a dependency-bearing representation realise its advantage on small models. A4 likewise shows its largest loss at $p\geq2/3$ ($-2.98$~pp). Consistent with this interpretation, the A4--A2 gap on the $p\geq2/3$ subset shrinks from 1.63~pp with Config-27B to 0.08~pp with Config-Plus (Appendix~\ref{app:scale-ablation}).

\noindent\textbf{Deviation feedback (A1).}
Removing the deviation signal $\delta$ raises search calls to $1.22\times$, lowers the within-budget success rate to 86\%, and causes the largest Overall drop among the single ablations ($1.96$~pp). The planner still perceives that information is missing but cannot identify the type of mismatch or select a matching escape strategy; it can only append homogeneous searches until budget is exhausted. If implicit LLM judgement could substitute for $\delta$, removing it should not simultaneously inflate search volume and depress quality. In contrast, A1 does not become more damaging as $p$ rises.

\noindent\textbf{A3's coverage beyond accommodation.}
Among queries with $p=0$, A3 still loses $1.20$~pp across 74 scored queries, with search dispatch contracting to $0.41\times$. Combined with the analysis above, this indicates that interpretive update is not only the cognitive basis for restructuring but also the baseline maintenance mechanism for mental-model evolution as a whole: every query's plan evolution depends on the coherent state that it sustains. This is the empirical grounding for treating it as a foundational loop in \S\ref{sec:intro} rather than a localised mechanism.

\noindent\textbf{A\textsubscript{full}.}
A\textsubscript{full}'s search ratio of $0.54\times$ sits between A1 ($1.22\times$) and A3 ($0.41\times$), reflecting a mixture of two failure modes: it inherits A3's contracted external search, partially offset by A1's tendency to over-append. At $p\geq2/3$, $-$4.73~pp matches the severe A2/A3 range; at $p=0$, $-$2.14~pp reflects the compounded broad-coverage failure of interpretive update and deviation feedback.

Overall, the matched losses on the $p\geq2/3$ subset are statistically significant for A2, A3, A4, and A\textsubscript{full} (all $p_{\mathrm W}\leq .014$), but not for A1 ($p_{\mathrm W}=.496$). This pattern is consistent with the analysis above.

\section{Discussion}
\label{sec:discussion}
\textbf{Insight consistency: how the loops drive the mental model toward reality.}
Insight measures the quality of integrative reasoning, and our experiments show this dimension does depend on reasoning power (\autoref{tab:leaderboard}). Yet with small backbones, VeriTrace's Insight still surpasses baselines that use Claude-Sonnet-4.5 and other frontier backbones. Reasoning power matters, but it is clearly not the only determinant of Insight, or this would not happen. The difference lies not only in how strong the model is when it reasons, but in what it reasons over. Even a strong backbone, integrating over a mental model polluted by mixed information and noise, and misaligned in its frame, produces analysis built on distorted information. The three loops do not change single-pass reasoning; across the search process they give the agent three capacities: (1) critical learning, monitoring the current understanding to judge which new evidence the current frame needs and how to fold it in; (2) deviation awareness, typing each search deviation so the system can distinguish a strategic failure from a framing failure; and (3) frame rebuilding, restructuring the frame once evidence shows it is wrong so it better holds incoming information. These three capacities let the agent maintain a fact-consistent mental model whose frame is not derailed, so that the final writing and reasoning rest on fact-consistent logic and yield valuable judgments and recommendations. The per-dimension ablations show that removing all four mechanisms reduces Insight by 4.35~pp, the largest drop across the four dimensions, and nearly removes VeriTrace's matched-backbone Insight advantage (Appendix~\ref{app:ablation-perdim}). This pattern is consistent with the interpretation above.

\textbf{Scale and regulation create headroom along different axes.}
We do not claim that scale is unimportant, or that regulation replaces scale, but that part of the apparent scale dependence in deep research stems from missing regulatory loops. Within VeriTrace, scaling the backbone yields modest Overall gains. Under a fixed backbone (\autoref{tab:main}), however, VeriTrace exceeds the strongest matched baseline by clear margins on both Overall and Insight, and removing the regulatory loops (A\textsubscript{full}, \autoref{tab:ablation}) drops the system below that baseline; with other components held fixed, this excludes attributing the gain to writer quality or prompt style. The same holds on Overall: VeriTrace produces the strongest reproducible open-source result in \autoref{tab:leaderboard}, exceeding several systems that use frontier closed-source backbones.

\section{Conclusion}
\label{sec:conclusion}

We presented VeriTrace, a cognitive-graph implementation that regulates a deep research agent's mental model through three feedback loops, namely interpretive update, deviation feedback, and schema revision, steering the model toward what reality supports rather than treating persistent artefacts as larger memory stores. Controlled comparisons and ablations show that explicit regulatory loops improve performance beyond backbone scale. More broadly, VeriTrace operationalises self-regulation as an executable mechanism rather than a cognitive-science metaphor, complementary to scaling. As agents take on longer-horizon, more uncertain tasks beyond deep research, we argue that the maturity of their self-regulation should become a primary lens for evaluating and diagnosing them.
\section*{Limitations}

(1)~Evaluation centres on DeepResearch Bench and DeepConsult; generalisability to other open-environment tasks such as web navigation, scientific discovery, and exploratory debugging remains to be validated. (2)~Restructuring triggers are heuristic; we provide no formal optimality guarantees. (3)~The cognitive graph is built from scratch per task with no cross-task transfer. (4)~Experiments use a limited set of non-frontier Qwen and DeepSeek configurations; broader validation across additional model families remains future work.

\section*{Ethical Considerations}

Deep research agents raise four practical concerns. First, they retrieve and quote public web content. We use only short excerpts linked to their sources, follow robots-exclusion rules, and do not access content behind paywalls or logins. Second, web retrieval can return unreliable or disputed sources. VeriTrace reduces this risk by scoring source quality and recording disagreements across sources. It cannot fully prevent citations from drifting away from their sources or bias from early search results. These outputs remain agent-assisted drafts and should therefore be checked by people before use in medicine, law, finance, or other high-stakes settings. Third, deep research requires more computation than a single model response. Our design limits this cost but does not remove it, so we recommend running jobs in batches on shared hardware where possible. Finally, the evaluation benchmarks contain no personal identifiers. Production deployments that process user-provided queries should nevertheless follow standard privacy-preserving practices.

\bibliography{references}

\appendix
\renewcommand{\thesection}{\Roman{section}}

\section{Background: LLM-Based Multi-Agent Decision Making}
\label{app:prelim}

\paragraph{LLM multi-agent systems.}
LLM multi-agent systems (LLM-MAS) coordinate role-specialised language-model agents through explicit protocols rather than a single end-to-end chain-of-thought. Cooperative frameworks such as AutoGen \citep{wu2023autogen} and MetaGPT \citep{hong2024metagpt} show that exchanging structured artefacts between roles can outperform monolithic prompting on complex tasks. More broadly, recent LLM-agent studies have explored how agents can plan, communicate, and act over structured intermediate states, extending language-model reasoning beyond purely textual generation \citep{guo2024llmmas,li2023camel,qian2024chatdev}.

Beyond cooperative task-solving, some work treats agent decisions as control over structured non-textual variables \citep{long2026emodistillofflineemotionskill}. For example, agent behaviour can be parameterised by latent strategic states rather than only by natural-language instructions \citep{xie2021learning}. 
This perspective is related to VeriTrace's use of an explicit cognitive graph, although our focus is evidence organisation rather than negotiation policy learning.

Most current deep research agents \citep{du2025deepresearch,huang2025deep}, however, are still driven by a single planner that calls tools sequentially and relies on the LLM context to integrate evidence across rounds. In high-stakes negotiation, \citet{long2025emomas} instead study Bayesian orchestration over multiple agents, showing how explicit coordination mechanisms can support risk-sensitive decision-making. VeriTrace follows the broader principle of explicit coordination, but specialises it for verifiable research traces and evidence-state restructuring.

\paragraph{Structured intermediate states and targeted adaptation.}
Recent work across language and multimodal reasoning increasingly suggests that reliable reasoning benefits from explicitly structuring intermediate information rather than relying only on a monolithic model context.
For long-document agents, AWM introduces an answerable working-memory mechanism that selectively maintains information useful for downstream reasoning~\citep{zhou2026awmanswerableworkingmemory}.
A related trend appears in vision and multimodal models, where robustness and adaptation are improved by explicitly controlling which information, layers, or parameter directions are modified.
Counterfactual visual alignment has been used to mitigate visual laziness in multimodal LLMs~\citep{xiao2026stayingvigilantmitigatingvisual}, while layer-specific prompt fusion~\citep{xiao2026layerspecificpromptfusiondiscovery} and instance-aware prompt tuning~\citep{xiao2025visualinstanceawareprompttuning} adapt representations selectively rather than applying a single global prompt.
At the parameter level, Model Shapley shows that individual model parameters contribute unequally and interact cooperatively, and uses Shapley-based attribution to identify task-relevant parameter subsets for targeted fine-tuning and model intervention~\citep{NEURIPS2025_6724eae9}.
Recent work on structured, task-aware low-rank adaptation similarly argues that not all parameter directions contribute equally to downstream behaviour~\citep{xiao2026directionsmatterstructuredtaskaware}, consistent with the broader prompt-adaptation literature~\citep{xiao2026promptbasedadaptationlargescalevision}.
Although these studies target different modalities, they share a common principle with \textsc{VeriTrace}: complex reasoning can become more reliable when the system exposes and selectively updates structured intermediate states instead of leaving all evidence integration implicit inside the model context.

\paragraph{Multi-agent decision formulation.}
We assume a small set of role-specialised agents---a planner, $K$ parallel searchers, a reader, and a manager---that coordinate through a single shared, structured state rather than direct message passing among workers \citep{wu2023autogen,hong2024metagpt,guo2024llmmas}. 
At each step, only the planner makes high-level decisions, and these decisions are \emph{action-typed}: the planner chooses between dispatching evidence-gathering actions and proposing operations that revise the shared state itself. This follows the planner/worker decomposition that has proven effective for long-horizon LLM-agent tasks \citep{li2023camel,qian2024chatdev}.
Workers execute the dispatched actions independently; the manager folds their observations back into the shared state via a deterministic update operator. Related work has explored debt negotiation policies for multi-agent decision-making \citep{long2025emodebt}. In contrast, our setting uses structured shared states to support evidence verification rather than credit negotiation.

\section{Cognitive Graph Data Model}
\label{app:cg_schema}

Section~\ref{sec:cg} defines the cognitive graph abstractly as
$G_t = (S_t, C_t)$ over node and edge content. This appendix specifies the
concrete fields each node and edge carries, since several method-level claims
(cross-node routing in \S\ref{sec:assimilation}, axis rotation in
\S\ref{sec:accommodation}, evidence linking in \S\ref{sec:writing}) depend on
the layout below.

\paragraph{Concept Node.}
A Concept node $n_i$ groups its fields into six slots:

\textbf{Identity \& topology.} A UUID \emph{id}; a human-readable
\emph{name} (placeholders such as ``[supplier]'' allowed before discovery);
\emph{node\_type} $\in$ \{\emph{start}, \emph{placeholder},
\emph{discovered}\}; \emph{hop\_distance} (BFS layer from the
user-question root); \emph{discovery\_dependency} (upstream node IDs whose
findings must be known before this node's edges become searchable).

\textbf{Constraints.} \emph{type\_constraint} (e.g., ``a company'') and
\emph{condition\_constraints} (a list of filtering predicates derived from
the user query); these are static throughout a run.

\textbf{Acceptance criteria.} Criteria attached to tasks targeting $n_i$
come in two priority classes: \emph{core\_criteria} (required for
\textsc{known}) and \emph{supplementary\_criteria} (feed the quality
profile only). They accumulate on $n_i$ as $a_i$ in
\S\ref{sec:assimilation}; the core-priority subset is $a_i^\textup{core}$.
The manager reconciles findings against both classes and stores the
residues as \emph{core\_pending} and \emph{supplementary\_pending}
(Appendix~\ref{app:update}); the node flips to \textsc{known} when
\emph{core\_pending} becomes empty.

\textbf{Findings (the items mechanism).} Findings are stored as a two-level
structure rather than a flat list. \emph{discovered\_items} is the set of
concrete entities found for this node (e.g., \{Scale~AI, Appen, \ldots\});
\emph{item\_findings} indexes per-item attributes
(\{Scale~AI: \{revenue\_2024: ``\$1B [[3]]'', \ldots\}\});
\emph{cross\_item\_findings} stores patterns spanning items (e.g.,
\{market\_consolidation: ``top three firms control 60\% [[1]]''\}). The
(item, attribute) decomposition is what enables axis rotation by
$R_\textup{pivot}$ (\S\ref{sec:accommodation}).

\textbf{Quality and deviation tracking.} \emph{quality\_profile} aggregates
per-page CR-AAP into the two means $(\overline{\textup{cr}}_j^t,
\overline{\textup{aap}}_j^t)$ plus three categorical labels:
\emph{finding\_strength} $\in$ \{\emph{strong}, \emph{moderate},
\emph{weak}, \emph{none}\} (coverage of \emph{core\_pending});
\emph{unexpected\_strength} (the categorical $\psi_j^t$ used by the
strategy router; Appendix~\ref{app:reader}); and
\emph{accessibility\_barriers}, a list of barrier types
(\emph{paywall}, \emph{login\_required}, \emph{requires\_download},
\emph{dynamic\_load}). $\phi_j^t$ in Definition~\ref{def:deviation} is the
indicator of this list being non-empty.

\textbf{Audit trail.} \emph{contradictions} (typed five-tuples;
Appendix~\ref{app:update}); \emph{unexpected\_discoveries} (free-form,
criterion-divergent findings from Reader Part~B); \emph{temporal\_notes}
(a single-string aggregation of time-sensitive context across the node's
findings, e.g., ``Scale~AI terminated June 2024; data as of Dec 2023'',
distinct from the page-level \emph{temporal\_context} emitted by the
reader); \emph{related\_tasks} (the IDs of tasks whose findings touched
this node, used by the writer to fetch the right evidence subset
$\mathcal{E}_k$); \emph{cited\_refs} (global ref indices appearing in
the node's findings).

\paragraph{RelationEdge.}
An edge $e_{ij}$ holds an \emph{inquiry\_goal} (a self-contained search
question), per-task acceptance criteria \emph{core\_criteria} and
\emph{supplementary\_criteria} (instantiated per attempt as $b_{ij}^t$),
\emph{task\_type} $\in$ \{\emph{open}, \emph{specified}\} with optional
\emph{specified\_source} when the user named one, an \emph{attempt\_count}
$k_{ij}^t$ capped at $K\!=\!3$, and a \emph{state} $u_{ij}^t \in$
\{\emph{to-solve}, \emph{investigating}, \emph{solved},
\emph{exhausted}\}. Each attempt appends a \emph{search\_history} entry
(query, summary, planner feedback) so the planner can read past attempts
before issuing a strategy choice.

\paragraph{State machine.}
A node's \emph{cognitive\_state} is determined entirely by the assimilation
update of \S\ref{sec:assimilation}:
\emph{unknown} when no task has touched it, \emph{partial} once any finding
is written, \emph{known} once \emph{core\_pending} is empty.
\emph{start} is reserved for user-anchored entities that bypass search.
An edge's \emph{state} follows: \emph{to-solve} until first dispatch,
\emph{investigating} between attempts, \emph{solved} once
$a_j^\textup{core} \subseteq \textup{sat}(f_j^t)$, \emph{exhausted} when
$k_{ij}^t = K$ without convergence, at which point the planner is required
to switch strategy or propose restructuring (\S\ref{sec:dynamics}).

\section{Cognitive Graph Update Procedure}
\label{app:update}
\label{app:cg_update}

The cognitive graph manager realises the operator $\Gamma$ of
\S\ref{sec:assimilation} once per completed search task. It performs four
steps in sequence: \emph{extract} discovered items from the searcher's
response, \emph{route} each finding to the appropriate finding container,
\emph{reconcile} the findings against the task's acceptance criteria, and
\emph{branch} divergent material into the contradiction, unexpected, or
search-experience side channels. The procedure operates on the target
node $n_j$ of edge $e_{ij}$ and is driven by a single LLM call using the
update prompt (Appendix~\ref{app:prompts}, Cognitive Graph Manager update block, item~3).

\paragraph{Step 1: extract discovered items.}
A \emph{discovered item} is an entity that directly answers the node's
inquiry goal: an organisation, product, system, or distinct option. The
manager builds the \emph{discovered\_items} list from the searcher's
response, excluding metadata that does not stand as a research target on
its own (people names, dates, events, locations, raw metrics).
Rankings, top-N lists, and ordered enumerations discovered during search
are stored as ordered arrays in \emph{cross\_item\_findings} so that
the ranking context survives subsequent rounds.

\paragraph{Step 2: route findings by attribution.}
Every finding is routed by an attribution test. \emph{If the finding can
be attributed to one specific entry of \emph{discovered\_items}}, it
enters \emph{item\_findings} as a per-item attribute keyed by the item
name (e.g., \emph{\{Scale~AI:\,\{market\_position:\,...\}\}}).
\emph{Otherwise}, when it describes a pattern, trend, ranking, or
insight that spans items, it enters \emph{cross\_item\_findings},
either as a single-sentence string or as an ordered array. Each value
carries $[[m]]$ citations to the source records in $\mathcal{E}$. Empty
findings are also recorded: a confirmed negative result (e.g., ``no
public record of this partnership'') is itself a finding worth keeping,
since it removes the criterion from $\textit{pending}$ for the planner.

\paragraph{Step 3: reconcile against acceptance criteria.}
The manager compares the aggregate findings against the edge's
\emph{core\_criteria} and \emph{supplementary\_criteria}. The
residue, i.e., criteria not yet covered by any finding, is written back
as \emph{core\_pending} and \emph{supplementary\_pending} on $n_j$.
\textbf{The pending lists are residues, not copies}: an empty
\emph{core\_pending} is precisely the condition $a_j^\textup{core}
\subseteq \textup{sat}(f_j^t)$ that flips $n_j$
to \emph{known}. Coverage is judged at the criterion level rather
than by labelling each individual finding, since one criterion may
require several findings together (e.g., a market-share criterion may
need both a total figure and per-firm shares before it counts as
covered).

\paragraph{Step 4: branch divergent material.}
Findings that fit neither \emph{item\_findings} nor \emph{cross\_item\_findings} are routed into one of three side channels.

\textbf{Contradictions.} When a new finding clashes with an existing one on the same criterion, the manager appends a structured record to $n_j.\textit{contradictions}$ with fields $\{\textit{criterion},\allowbreak \textit{old\_claim},\allowbreak \textit{new\_claim},\allowbreak \textit{resolution},\allowbreak \textit{kept}\}$, where \textit{resolution} is a free-form rationale stating why one claim was retained (typically citing authority, recency, or source type) and \textit{kept} $\in \{\textit{old}, \textit{new}\}$ records which claim survives in the active findings. The recorded contradictions are surfaced to the planner in the next round as candidate triggers for $R_\textup{correct}$ (\S\ref{sec:accommodation}).

\textbf{Unexpected discoveries.} A finding is admitted to
\emph{unexpected\_discoveries} only if (i) it would help address the
dimension of the original question that the inquiry goal targets, and
(ii) it lies completely outside every entry of the acceptance criteria.
Both conditions must hold; condition~(i) excludes off-topic noise,
condition~(ii) excludes Part-A material the reader simply forgot to
attribute. Each entry is a one-sentence description with $[[m]]$
citations, ranked by contribution to the inquiry goal.

\textbf{Search experience.} A separate channel records the inability to
access information rather than the information itself. The manager
populates \emph{search\_experience} from two sources only: access
failures reported by the searcher (e.g., HTTP errors, CAPTCHA, timeout)
and confirmed data absence (the searched material verifiably does not
exist or is not publicly available). The mutual-exclusion rule is
explicit in the prompt: \emph{any finding that conveys actual
information (even partial, dated, or tangential) belongs to
\emph{item\_findings} or \emph{cross\_item\_findings}, never to
search\_experience}. The reader's \emph{accessibility\_notes}
(Appendix~\ref{app:reader}) feed both \emph{search\_experience} and
the node-level \emph{accessibility\_barriers} flag in
\emph{quality\_profile}.

\paragraph{Cross-node routing.}
The default and overwhelmingly common case is a single entry on the
planner-specified target node $n_j$. The manager additionally admits
an update on a different node $n_k$ ($k \neq j$) when a finding
satisfies one of $n_k$'s own acceptance criteria in $a_k$ rather than
$n_j$'s; the finding is written to $n_k$, and the originating task is
recorded in $n_k$'s associated task set,
$\mathcal{T}_k^{t+1} = \mathcal{T}_k^t \cup \{\textup{task}_t\}$, so
the writing pipeline can later trace which task supplied each node's
evidence. \emph{start} nodes are immutable search anchors and are
never included in \emph{cognitive\_updates}; information about a
\emph{start} entity is recorded on the relevant exploration node
instead.

\paragraph{Planner view (compiler).}
The planner does not consume $G_t$ directly. A graph compiler renders
$G_t$ into a Markdown planning context: per-node summaries (state,
constraints, findings, quality profile, accessibility barriers,
contradictions, recent unexpected discoveries) and per-edge sections
grouped by status (solved, investigating, exhausted, to-solve), with
search history shown for investigating edges. Verbatim evidence quotes
are not included; the planner sees abstracted findings and quality
signals only. This bounded rendering caps the planner's prefill at
$O(|V_\textup{frontier}|)$ regardless of total graph size.

\medskip
\noindent An excerpt from query~14 after the first search task completes
(translated for typographic clarity; structure preserved from the system's Markdown rendering; emoji omitted
and replaced by bracketed labels):

\begin{Verbatim}[fontsize=\scriptsize, frame=single, framesep=5pt,
                 rulecolor=\color{gmemBorder}, fillcolor=\color{gmemBg},
                 breaklines=true, breakanywhere=true]
## Current Cognitive Graph State
**Structure type**: mixed

### Entity Nodes
- [START] Math x Quantum crossover  (e1, hop=0)
- [KNOWN] Main research teams       (e2, hop=1)
  - Discovered: MIT CTP-LI, Stanford QFARM, Waterloo IQC,
    Oxford QG, ETH QIT, BIMSA, ... [17 total]
  - Cross-item insights:
    - geographic distribution [[5]][[12]]
    - math-tool innovation: Quon, ZX-calculus [[18]]
  - Supplementary pending: team_size_and_history
  - Unexpected (2):
    - "BIMSA Quon language [[36]]" ...
  - Quality: CR=3.9 AAP=4.3 | finding=strong, unexpected=moderate

### Resolved inquiry goals
- e1 --has_main_teams--> e2

### To-solve inquiry goals
- e2 --research_directions--> [...]
- e2 --funding_sources--> [...]
- e2 --industry_partners--> [...]
\end{Verbatim}

\section{Restructuring Operations and Safety Protocol}
\label{app:safety}

Section~\ref{sec:accommodation} defines the abstract operator family
$\mathcal{A}_\textup{struct} = \{R_\textup{conc}, R_\textup{aug},
R_\textup{pivot}, R_\textup{prune}, R_\textup{correct}\}$. This appendix
specifies (i)~the two-stage architecture that realises these operators,
(ii)~the primitive edit alphabet they compile to, (iii)~one concrete
realisation example per operator, and (iv)~the multi-phase safety protocol
under which edits are applied.

\paragraph{Two-stage architecture.}
The planner does not directly edit the graph. It emits an intent
$\rho_t = (k, \textit{rationale})$ via $\textsc{propose\_restructure}$, where
$k \in \mathcal{A}_\textup{struct}$ names the operator class and
\textit{rationale} is a free-form justification grounded in $\delta_j^t$ and
the contradictions log. The cognitive graph manager then translates this intent into a concrete sequence over the
primitive edit alphabet

\begin{equation}
\begin{split}
  \Sigma = \{\,&\textsc{add\_node},\ \textsc{add\_edge},\\
               &\textsc{remove\_node},\ \textsc{modify\_node},\\
               &\textsc{modify\_edge}\,\}.
\end{split}
\end{equation}
There is no explicit \emph{remove\_edge}. An edge is removed only as a
side effect of \textsc{remove\_node}, which deletes the node together
with all of its incident edges. Omitting a standalone edge deletion is
deliberate: removing a node's sole incoming edge would detach it from
the user-question root, and Phase~2 reconnection would then restore
reachability by re-adding a bridging edge, so a standalone edge
deletion is cancelled by the architecture rather than realised.

\paragraph{Operator semantics and example realisations.}
The five abstract operators each compile to a characteristic
$\Sigma$-pattern.

\textbf{$R_\textup{conc}$ (concretisation).} Triggered when a single concept
node accumulates a list of \emph{discovered\_items} that downstream edges
must investigate per-item. Realisation: $m$~\textsc{add\_node} for the
children, $m$~\textsc{add\_edge} from the parent, and rerouting of any
pre-existing downstream edges to the appropriate child.
\emph{Example.}\linebreak[1] The ``Technology Routes'' node has discovered items
\{RL, end-to-end, modular, sim-to-real\} and expands into four child nodes,
each inheriting the downstream ``maturity assessment'' edge.

\textbf{$R_\textup{aug}$ (augmentation).} Triggered when Part-B insights or
contradictions reveal a question dimension absent from the initial graph.
Realisation: \textsc{add\_node} (the new dimension) plus \textsc{add\_edge}
(wiring it into the relevant subgraph). A candidate dimension is admitted
only if it materially helps answer the original question, not merely a
tangential topic.

\textbf{$R_\textup{pivot}$ (axis rotation).} Triggered when the cognitive
graph contains an $N \!\times\! M$ structure ($N$ items $\times$ $M$
dimension nodes) but the searched evidence reveals that information
co-locates by item, not by dimension: a single source per item covers all
$M$ dimensions.

Realisation: $N$~\textsc{add\_node} (one per item, with the
$M$ dimensions absorbed into the new node's \emph{core\_criteria}),
$N$~\textsc{add\_edge} (parent to each item), and $M$~\textsc{remove\_node}
(the dimension nodes whose content has migrated).

\emph{Example.} ``Semiconductor Product Lines'' with downstream dimension
nodes \{tech advantages, market share, supply-chain risk\} pivots into
four product-line nodes (mobile / server / AI / memory chips), each with
one inquiry that bundles all three dimensions; the dimension nodes are
removed. Net effect: $4 \!\times\! 3 = 12$ tasks collapse to $4$.

\textbf{$R_\textup{prune}$ (pruning).} Triggered when search has
established that a future node is irrelevant to the original question.
Realisation: \textsc{remove\_node}, with cascading deletion of incident
edges. The manager refuses the proposal if (a) the node is in
$N_\textup{user}$ (\textbf{I2}), (b) the node has non-empty findings
$f_i^t \neq \emptyset$ (\textbf{I1}), or (c) the rationale cites local
search difficulty rather than structural irrelevance.

\textbf{$R_\textup{correct}$ (premise correction).} Triggered when an open
contradiction at $n_i$ invalidates a premise on which downstream nodes
were constructed. Realisation: \textsc{modify\_edge} on each $e_{kj}$ for
$n_j \in \downarrow\!n_i$ (rewriting $g_{kj}$) and \textsc{modify\_node} on
each affected $n_j$ (rewriting $a_j$). $f_j^t$ is left untouched by these
primitives, satisfying \textbf{I1}.

\emph{Example.} A node investigating firm X's strategy reveals that
X was acquired by Y in late 2024, contradicting the premise of
independent operation on which several downstream nodes (X's
partnership pipeline, capital structure) were constructed.
$R_\textup{correct}$ rewrites those downstream inquiry goals around
X's role within Y's portfolio while preserving the acquisition-discovery
finding on $n_i$.

\paragraph{Phase 1: surgical edit.}
The manager receives $\rho_t$, $S_t$, and the protected sets
($N_\textup{user}$ and any node with $f_i^t \!\neq\! \emptyset$),
and emits a single $\Sigma$-sequence. Any \textsc{remove\_node}
targeting a node in $N_\textup{user}$ is refused, enforcing
\textbf{I2}. On refusal, $S_{t+1} = S_t$ and the rejection reason
is returned to the planner.

\paragraph{Phase 2: orphan reconnection.}
A Phase-1 edit may disconnect nodes from the user-question root by removing
intermediate dimensions. The manager runs up to five reconnection rounds,
each adding bridging edges to restore reachability while respecting the
protected sets. If reachability is not restored within five rounds, the
system rolls back to the snapshot $\widetilde{S}_t$ taken before Phase~1.

\paragraph{Invariants in execution.}
\textbf{I1} holds for the evidence stored in $\mathcal{E}$, which has no
deletion path. The cognitive-graph attribution layer ($f_j^t$) is also
append-only under all primitives except one: \textsc{modify\_node} admits
a \emph{remove\_items} parameter that lets the planner discard the
per-item attribution of a discovered entity it has reclassified as not
a research target (e.g., a noise entry inadvertently recorded).

The verbatim quotes in $\mathcal{E}$ are unaffected by this revision; only the (item, attribute) mapping in \emph{item\_findings} is removed.
\textsc{remove\_node} requires $f_i^t = \emptyset$ for $R_\textup{prune}$
and is refused by Phase~1 on any node carrying findings; the remaining
primitives only add. \textbf{I2} holds because Phase~1 catches
$N_\textup{user}$ deletions before they commit.

\section{Reader Output and Unexpected Findings}
\label{app:reader}

The reader runs once per page selected by the searcher and produces two
streams that feed different downstream consumers. Part~A populates
$f_j^t$; Part~B populates the divergence signal $\psi_j^t$ used by the
strategy router (\S\ref{sec:strategy}). Splitting them prevents Part~A's
prior expectations from filtering out genuinely unexpected content.

\paragraph{Part A: criterion-aligned findings.}
For each entry in the per-task acceptance criteria $b_{ij}^t$, the reader
returns either a finding or marks it as a gap. A finding has three required
fields: a one-sentence answer, a 200--500 character verbatim quote from the
page, and the source URL. The verbatim quote is the load-bearing constraint:
findings without an exact-string quote that supports the answer are dropped
before they reach the manager. This is what makes each (answer, evidence)
pair a citation-traceable record in the evidence store $\mathcal{E}$
(\S\ref{sec:writing}). A criterion is \emph{satisfied} only when the
evidence contains values that can be cited verbatim in a final report
(numbers, named entities, dated facts), not merely metadata describing
where data exists.

\paragraph{Part B: divergence track.}
Independently of Part~A, the reader is asked what the page reveals
\emph{beyond} $b_{ij}^t$. The output is a single field
\emph{unexpected\_insights}, a list of $\{\emph{label}, \emph{insight},
\emph{evidence}\}$ entries. The prompt directs the reader to look for
four kinds of content under this field: facts that contradict or nuance
the planner's assumptions, important entities or relationships not asked
about, timeline events that change the picture, and leads worth
investigating further. These four are guidance for what counts as
unexpected; they are not separate output fields.

\paragraph{Defining ``unexpected''.}
A Part-B insight qualifies as unexpected only if both conditions hold:
(i) it would help answer the dimension of the original question that this
task targets, and (ii) it lies outside every entry of $b_{ij}^t$
\emph{and} every active entry of $a_k$ for the immediate neighbours of
$n_j$ in $G_t$. Condition~(i) is what distinguishes ``unexpected'' from
``off-topic noise''; condition~(ii) is what distinguishes ``unexpected''
from Part~A material the reader simply forgot to attribute. Failures of~(i)
are dropped; failures of~(ii) are reclassified into Part~A by the manager.

\paragraph{From Part B to $\psi_j^t$.}
The manager aggregates Part-B insights across all pages of the current
task into a single ordinal $\psi_j^t \in$ \{\emph{none}, \emph{weak},
\emph{moderate}, \emph{strong}\}, scored from the joint volume and
topical coherence of the insights (rather than from any single page). This
ordinal is the categorical signal consumed by the \textsc{Unex} predicate
(Eq.~\ref{eq:strategy}); the [1,4] interval in
Definition~\ref{def:deviation} is a linear projection used for notational
uniformity with CR/AAP. The threshold $\tau_\psi$ corresponds to ``moderate
or stronger'' on this ordinal scale (Appendix~\ref{app:strategies}).

\paragraph{Accessibility notes.}
The reader also emits a binary accessibility indicator $\phi_m \in
\{0,1\}$ with a reason tag for each page; the node-level flag
$\phi_j^t = \max_m \phi_m$ is the third component of $\delta_j^t$.
Detection rubric, exclusion rule from CR-AAP aggregation, and routing
implications are in Appendix~\ref{app:craap}.

\section{Search Strategy Execution}
\label{app:strategies}

Section~\ref{sec:strategy} defines the strategy router
$\pi : \mathcal{D} \to \mathcal{A}_\textup{search}$ as a priority-ordered case
mapping. This appendix specifies (i) the threshold values that instantiate
the predicates, (ii) the operational semantics of each branch, and (iii) the
form of the planner's next $\textsc{add\_task}$ call once a strategy is
chosen.

\paragraph{Thresholds.}
$\tau_h = 3.5$ (mid-Likert, between ``adequate'' and ``good'') instantiates
both $\textsc{Rel}$ and $\textsc{Cred}$. $\tau_\psi$ corresponds to
``moderate or stronger'' on the four-level
\emph{unexpected\_strength} ordinal (Appendix~\ref{app:reader}).
$\textsc{Accessible}(\delta) \!\equiv\! (|\emph{accessibility\_barriers}|
\!=\! 0)$ uses the indicator $\phi_j^t$ directly. An auxiliary low-quality
threshold $\tau_l = 2.5$ governs structural escalation: when
$\overline{\textup{cr}}_j^t \!<\! \tau_l$ \emph{and}
$\overline{\textup{aap}}_j^t \!<\! \tau_l$, the manager raises a
quality-gap signal that the planner consumes as input to
$\mathcal{A}_\textup{struct}$ (\S\ref{sec:accommodation}).

\paragraph{Branch semantics.}

\textbf{\emph{exploit}} ($\textsc{Rel} \wedge \textsc{Cred}$). The current
source family is delivering. The next $\textsc{add\_task}$ keeps source type
and language fixed and narrows the inquiry goal to a sub-criterion of
$a_j^\textup{core}$ that the prior pages partially answered, typically a
deeper attribute of an item already in
\emph{discovered\_items}.

\textbf{\emph{verify}} ($\textsc{Rel} \wedge \neg\textsc{Cred}$). Content
fits but authority is weak. The next $\textsc{add\_task}$ targets the same
criterion in a different source family (primary documents over news,
peer-reviewed over blog, official over secondary). One successful
cross-source attempt is sufficient; over-verification is explicitly
discouraged. If a prior \emph{verify} on the same edge also yielded
$\overline{\textup{aap}} \!<\! \tau_h$, the planner accepts high-CR sources
and triangulates by source count instead, since the domain's authority
ceiling is intrinsically low.

\textbf{\emph{substitute}} ($\neg\textsc{Accessible}$). Data exists but is
gated. The next $\textsc{add\_task}$ relaxes both source type and authority:
paywalled paper $\to$ press release; closed site $\to$ open mirror; native
document $\to$ analyst summary. The planner is instructed to suggest source
\emph{directions} (e.g., ``industry-analysis blogs'', ``news deep-dives'')
rather than lock to a specific URL.

\textbf{\emph{pivot}} ($\neg\textsc{Rel} \wedge \textsc{Unex}$). The
original direction is barren but Part~B uncovered something topical. The
planner reads \emph{unexpected\_discoveries}, picks one entry, and either
(a) creates a new task on the same edge with the goal rewritten around the
unexpected angle, or (b) creates a new edge to a fresh node when the
discovery names a distinct entity. Case~(b) is the typical mechanism for
in-search graph growth.

\textbf{\emph{explore}} (default). Neither content fits nor a clear pivot
exists; the landscape is undermapped. The next $\textsc{add\_task}$ broadens
keywords, swaps query language (English $\leftrightarrow$ Chinese; up to
three languages on multilingual queries), and falls back to broader engines.
The planner is explicitly instructed to \emph{map} the landscape,
not guess a new direction; guessing is reserved for \emph{pivot}, which
has unexpected discoveries to ground it.

\paragraph{Re-evaluation cadence.}
$\pi(\delta_j^t)$ is recomputed at the close of each task on $e_{ij}$. The
strategy is therefore not a one-shot routing decision but the planner's
running diagnosis. Edges that reach $k_{ij} = K$ without ever escaping the
same deviation region (e.g., three consecutive \emph{explore} attempts that all
yield $\overline{\textup{cr}} \!<\! \tau_h$) are exactly the
structurally-misaligned edges that trigger
$\mathcal{A}_\textup{struct}$ proposals (\S\ref{sec:accommodation}).

\section{PreFilter Stage}
\label{app:prefilter}

\paragraph{Role.} PreFilter is a lightweight pre-screening stage that performs a binary classification on each raw search-engine result (title + snippet) before any full-page reading: \texttt{read} (send to the Reader for verbatim extraction) or \texttt{skip} (off-topic or duplicate). The classification is intentionally conservative: when unsure, the prompt instructs the model to classify as \texttt{read}, so PreFilter only prunes obviously irrelevant or redundant pages rather than acting as a relevance scorer. This keeps the Reader's context window from being saturated by clearly off-topic hits while preserving recall on borderline cases.

\paragraph{Per-configuration assignment.} PreFilter is deliberately assigned a lighter model than the Reader, since binary classification on title + snippet does not require deep reasoning (Table~\ref{tab:prefilter_assignment}).

\begin{table}[h]
\centering
\fontsize{7.5}{8.5}\selectfont
\setlength{\tabcolsep}{3pt}
\renewcommand{\arraystretch}{0.85}
\begin{tabular*}{\columnwidth}{@{\extracolsep{\fill}} lll @{}}
\toprule
\rowcolor{groupbg} \textbf{Config} & \textbf{PreFilter} & \textbf{Reader} \\
\midrule
Config-3B        & Q3.5-35B-A3B & Q3.5-35B-A3B \\
Config-27B       & Q3.5-35B-A3B & Q3.5-35B-A3B \\
Config-Plus      & Q3.5-35B-A3B & Q3.5-122B-A10B \\
Config-DeepSeek  & DSv4-Flash   & DSv4-Flash \\
\bottomrule
\end{tabular*}
\caption{PreFilter and Reader model assignments. In three configurations, PreFilter shares the Reader's lightweight backbone; in Config-Plus, PreFilter is dropped to Q3.5-35B-A3B while the Reader uses the heavier Q3.5-122B-A10B for cost efficiency.}
\label{tab:prefilter_assignment}
\end{table}

\paragraph{Why a lightweight model suffices.} Running PreFilter on a small, parallel-batched LLM sharply reduces the number of expensive read operations downstream: in Config-Plus, the Reader processes only ${\sim}148$ pages per query against ${\sim}580$ search hits returned (a ${\sim}74\%$ reduction; cf. Appendix~\ref{app:cost}), leaving the Reader free to spend its full token budget on the \texttt{read} subset. The Reader still performs the load-bearing verbatim-quote check on every page it sees (Appendix~\ref{app:reader}), so PreFilter's conservative bias does not propagate downstream errors. The full PreFilter prompt is reproduced as Agent~5 in Appendix~\ref{app:prompts}.

\section{Reader Scoring Mechanism}
\label{app:craap}

\begin{table*}[tp]
\centering
\footnotesize
\setlength{\tabcolsep}{5pt}
\renewcommand{\arraystretch}{1.15}
\begin{tabular}{@{}c l p{0.34\linewidth} p{0.34\linewidth}@{}}
\toprule
\textbf{Symbol} & \textbf{Dimension} & \textbf{Score 5 (anchor high)} &
\textbf{Score 1 (anchor low)} \\
\midrule
$c_m$       & Currency  & Within target temporal window (e.g.,
$\leq\!6$~months for time-sensitive queries; matched period for
historical queries).
& Wrong era; likely invalid due to age. \\[2pt]
$r_m$       & Relevance & Findings directly substantiate one or more
core criteria with citable values (numbers, dates, named entities);
usable verbatim in a final report.
& No substantive connection, or key data exists only behind an
inaccessible barrier (routed to \emph{accessibility\_notes}). \\[2pt]
$\alpha_m$  & Authority & Primary source, peer-reviewed venue,
official organisational publication, or recognised expert author.
& Anonymous, unattributable, or self-published with no editorial
gate. \\[2pt]
$\beta_m$   & Accuracy  & Claims verifiable, upstream citations
present, no factual errors against the cognitive graph.
& Unsupported assertions; sensational or contradicting known facts.
\\[2pt]
$\rho_m$    & Purpose   & Educational or analytical intent; objective
tone; no commercial interest.
& Agenda-driven; misleading framing; undisclosed conflicts. \\
\bottomrule
\end{tabular}
\caption{Per-page CR-AAP rubric used by the Reader. Both ends of each
dimension are anchored; intermediate scores degrade with
prompt-specified boundaries between adjacent levels. The five raw
ratings combine into the page-level composites
$\textup{cr}_m, \textup{aap}_m$ via
Eq.~\ref{eq:cr-page}--\ref{eq:aap-page}.}
\label{tab:craap_rubric}
\end{table*}

CR-AAP \citep{blakeslee2004craap} supplies two orthogonal evidence-quality
axes that we use to instantiate the page-level scores feeding both the node
profile $\bm{q}_j^t$ (Definition~\ref{def:node}) and the
relevance/credibility components of the deviation signal $\delta_j^t$
(Definition~\ref{def:deviation}). This appendix specifies the Reader's
per-page scoring rubric, the page-to-node aggregation, and the categorical
strength labels that the strategy router consumes.

\paragraph{Per-page Likert ratings.}
For each retrieved page $m$, the Reader produces five ratings in
$\{1,\dots,5\}$ from anchored rubrics (Table~\ref{tab:craap_rubric}):
currency $c_m$, relevance $r_m$, authority $\alpha_m$, accuracy $\beta_m$,
and purpose $\rho_m$. Anchors are explicit at both ends: score~$5$
requires the conditions in Table~\ref{tab:craap_rubric}; score~$1$ is the
inverse condition (out-of-date, off-topic, anonymous, unsourced,
promotional). Intermediate scores degrade in the Reader's prompt with
explicit boundary descriptions for each step rather than leaving the
midpoint unspecified.

\paragraph{Obtained vs.\ exists.}
The relevance dimension carries a load-bearing constraint that
distinguishes data the Reader \emph{extracted} from data the page merely
\emph{describes}. A page that says ``this platform provides 10 years of
market data'' but does not expose the values themselves is rated
$r_m \!\leq\! 2$, regardless of how relevant the topic is. The operational
test in the prompt is: \emph{can the findings on this page be cited
directly in a final report?} If the answer is no because the data are behind a
paywall or require a download, login, or dynamic load that the Reader could not
access, the information is recorded in \emph{accessibility\_notes}
(Appendix~\ref{app:reader}) and explicitly excluded from $r_m$. This
prevents a structural failure mode in which evidence appears
high-authority and on-topic on paper but is not actually available for the
writer to cite.

\paragraph{Composites.}
The two page-level composites are weighted sums

\begin{align}
  \textup{cr}_m   &= 0.30\, c_m + 0.70\, r_m, \label{eq:cr-page}\\
  \textup{aap}_m  &= 0.40\, \alpha_m + 0.35\, \beta_m + 0.25\, \rho_m,
                     \label{eq:aap-page}
\end{align}
so $\textup{cr}_m, \textup{aap}_m \in [1, 5]$. The relative weight
$w^{\textup{cr}}_R = 0.70$ encodes that, for research-style queries,
content alignment dominates recency when the two compete; the descending
$\bm{w}^{\textup{aap}}$ entries reflect that authority carries more signal
than internal accuracy or rhetorical purpose for the long-tail sources
typical of low-visibility research.

\paragraph{Page-to-node aggregation.}
Let $P_j^t$ denote the page set retrieved on $n_j$ up to time $t$, capped
at $|P_j^t| \!\leq\! 50$ by the rolling window of
Appendix~\ref{app:update}. The node-level entries of
$\bm{q}_j^t = (\overline{\textup{cr}}_j^t, \overline{\textup{aap}}_j^t)$
are per-page means:

{\small
\begin{align*}
\overline{\textup{cr}}_j^t  &= \frac{1}{|P_j^t|} \sum_{m \in P_j^t}
                                \textup{cr}_m, \\
\overline{\textup{aap}}_j^t &= \frac{1}{|P_j^t|} \sum_{m \in P_j^t}
                                \textup{aap}_m.
\end{align*}
}%
The means feed the $\textsc{Rel}$ and $\textsc{Cred}$ predicates of the
strategy router (Eq.~\ref{eq:strategy}); per-axis minima are tracked
alongside $\bm{q}_j^t$ to surface a single weak page in an otherwise strong
batch as a quality-gap signal.

\paragraph{Categorical strength labels.}
Two ordinal labels on $\bm{q}_j^t$ are produced by the cognitive graph
manager (not the Reader) at task closure, summarising the entire batch of
findings on $n_j$:
\begin{itemize}
  \item \emph{finding\_strength} $\in$ \{\emph{none}, \emph{weak},
        \emph{moderate}, \emph{strong}\} rates the substantive coverage
        of $n_j$'s pending criteria across the task's pages: multiple core
        criteria answered with cited evidence is \emph{strong}; thin
        evidence with major gaps is \emph{weak}.
  \item \emph{unexpected\_strength} (the categorical $\psi_j^t$ used by
        Eq.~\ref{eq:strategy}) rates the joint volume and topical
        coherence of Part-B insights aggregated across all pages of the
        task (Appendix~\ref{app:reader}): discoveries that open a
        plausible new direction relevant to the original question are
        \emph{strong}; minor observations are \emph{weak}.
\end{itemize}
These ordinals are LLM-judged at the batch level rather than thresholded
from per-page scores, because both ``substantive coverage'' and ``coherent
unexpected angle'' are properties of the finding set as a whole, not of
any single page.

\paragraph{Accessibility detection.}
For each retrieved page $m$, the Reader emits a binary indicator
$\phi_m \in \{0,1\}$ together with one of four reason tags
(\emph{paywall}, \emph{login\_required}, \emph{requires\_download},
\emph{dynamic\_load}). Detection is \emph{prompt-based}, not based on
HTTP status: the Reader infers the barrier from page text such as ``log in
to view'', ``subscription required'', or ``download the full dataset''.
HTTP-level detection would miss a substantial fraction of barriers, since
many gated pages return 200 with a teaser body. Inaccessible pages
contribute only $\phi_m$ and the reason tag, and are excluded from
Eq.~\ref{eq:cr-page}--\ref{eq:aap-page} so that an inaccessible page
cannot drag down the per-page composites of accessible pages in the same
batch. The node-level flag is the disjunction
$\phi_j^t = \max_{m \in P_j^t} \phi_m$, and the third component of
$\delta_j^t$ in Definition~\ref{def:deviation}; whenever $\phi_j^t = 1$,
the strategy router routes the next attempt on $e_{ij}$ to
\emph{substitute} (Eq.~\ref{eq:strategy}).

\paragraph{Why orthogonality matters.}
The two CR-AAP axes are scored and aggregated independently. This is what
enables differential diagnosis at the strategy router: a high-authority
but off-topic source reports as
$(\overline{\textup{cr}}_j^t \!<\! \tau_h,\;
  \overline{\textup{aap}}_j^t \!\geq\! \tau_h)$ and routes to
\emph{explore}, whereas a topic-aligned but low-credibility source reports as
$(\overline{\textup{cr}}_j^t \!\geq\! \tau_h,\;
  \overline{\textup{aap}}_j^t \!<\! \tau_h)$ and routes to
\emph{verify}. Collapsing the two axes into a scalar quality score would
render these two cases indistinguishable and force the planner back onto
undifferentiated retry.

\section{Writing Pipeline}
\label{app:writing}

The three-layer writing pipeline (\S\ref{sec:writing}) consumes the
terminal cognitive graph $G_T$ and the evidence store $\mathcal{E}$ and
produces the final report through three LLM stages whose information
access narrows in lockstep with their decision authority. This appendix
specifies the input, output, and information boundary of each layer,
and the two safeguards that make citations traceable.

\paragraph{Layer 1 --- outline planner.}
A single LLM call reads the cognitive graph (root anchor nodes without
findings are skipped) together with an
\emph{evidence-availability index} that lists, per node, the distinct
$m$ values present in $\mathcal{E}$ together with each record's
criterion tag. It produces a structured outline whose top level
captures the overall theme and an enumeration of user requirements,
and whose section list contains entries
$(\textit{section\_id},\allowbreak\, \textit{title},\allowbreak\, \textit{description},\allowbreak\,
  \textit{answers\_aspect},\allowbreak\, V_k)$, where $V_k$ is the
\emph{relevant\_node\_ids} subset whose findings the section covers
and \emph{answers\_aspect} records which aspect of the original user
question the section answers. The availability index is what prevents
proposing sections backed by too-thin evidence: if no node in a
candidate $V_k$ carries enough records, the section is dropped at this
layer rather than padded with prose later.

\paragraph{Layer 2 --- section planner.}
$S$ section planners run sequentially, one per section. Each planner
sees \emph{only} the findings of its bound subgraph $V_k$, an evidence
\emph{index} restricted to that section
$\{(m, \ell_m) : m \!\in\! \bigcup_{n_k \in V_k}\!\mathcal{E}_k\}$
(each $\ell_m$ already names an acceptance criterion of the section),
and the already-completed plans of earlier-numbered sections (with
their $[\![m]\!]$ markers stripped, so prior citations cannot leak as
inputs). Crucially, the planner does \emph{not} see the verbatim
quotes $v_m$ at this layer, only the index. The sequential ordering
is what enables cross-section deduplication: each section's prompt
instructs the planner to skip any insight semantically subsumed by a
prior section. Its output is a list of \emph{insights with bound
evidence}: each insight is a tuple
$(\textit{claim}, \mathcal{I}, \textit{hint})$ where
$\mathcal{I} \subseteq \mathcal{E}_k$, $|\mathcal{I}| \in [2, 5]$, and
\textit{hint} is an optional \emph{presentation\_hint} $\in$
\{\emph{table}, \emph{list}, \emph{comparison}, \emph{narrative}\}
that the writer may consult.

\paragraph{Layer 3 --- section writer.}
$S$ section writers run sequentially. Each writer receives the section
spec, the planned insights with their bound evidence IDs, and the
verbatim quotes $v_m$ for those IDs only; all other records in
$\mathcal{E}$ are filtered out before the prompt is constructed. The
writer also receives a tail of the previously written sections for
narrative continuity, capped by a configurable character budget
(default $15{,}000$ characters; per-language ceilings of $20{,}000$
for Chinese and $40{,}000$ for English when raised); the citation
markers in that tail are stripped, so the writer cannot re-cite
evidence from earlier sections. The writer renders prose with
$[\![m]\!]$ markers anchored to the bound evidence IDs.

\paragraph{Two citation safeguards.}
The pipeline guards against citation drift at two layers.
\emph{Information boundary:} the writer's evidence input is restricted
to the section's bound subset, so $[\![m]\!]$ values for records
outside that subset are unavailable at generation time.
\emph{Post-filter:} after generation, a deterministic regex filter
strips any $[\![m]\!]$ whose $m$ is not in
$\bigcup_{(c,\mathcal{I}) \in \textit{plan}_k}\!\mathcal{I}$, logging
each removal for audit. The information boundary is the primary
defence; the post-filter catches the rare case where a writer
hallucinates a marker outside the visible set.

\paragraph{Evidence store insertion.}
$\mathcal{E}$ is populated incrementally during the search loop, not
at write time. Each reader output yields one or more records; two
transformations are applied at insertion. \emph{Local-to-global
remapping} translates each reader's per-page index into the global
$m$-space used by the writer, so the same source page cited in
different tasks receives a single canonical $m$. \emph{URL
deduplication} collapses multiple records sharing the same
$\textup{src}_m$ into one entry while preserving the criterion-keyed
verbatim quote $v_m$: a single source page yields multiple records
only when the reader extracts evidence under several acceptance
criteria, with each record tagged by its own $\ell_m$. Records enter
$\mathcal{E}$ without further LLM processing; this mechanical
side-channel is what keeps the writer's $[\![m]\!]$ resolvable to a
verbatim quote at audit time.

\section{Termination Control}
\label{app:termination}

Termination is governed by the planner's \emph{finish()} action,
together with a code-level convergence guard and a wall-clock soft
deadline. The two conditions of \S\ref{sec:dynamics} describe the
ideal termination state; their enforcement is split across mechanical
and prompt layers.

\paragraph{Convergence guard.}
A call to \emph{finish()} signals the planner's intent to terminate,
but the system applies a guard before accepting it. The guard
mechanically enforces the node-level condition $\forall n_i \in N_T:\;
s_i^T \neq \textsc{unknown}$, i.e., every node must carry at least one
finding. The edge-level condition is enforced at the prompt layer
rather than the code layer: the planner is instructed to either
explore each remaining \emph{to-solve} edge or prune the node it
targets via \emph{propose\_restructure()} with $R_\textup{prune}$
before calling \emph{finish()}. In practice, stale \emph{to-solve}
edges typically target nodes that are themselves still
\emph{unknown}, and are therefore caught by the node-level guard
as a side effect. When the guard fires, the call is
rejected with a structured message listing each offending node (with
its incoming edges for context), and the planner is required to either
(i)~dispatch additional search tasks targeting these nodes, or
(ii)~call \emph{propose\_restructure()} with \emph{remove\_node}
($R_\textup{prune}$) for nodes whose accumulated evidence shows them
irrelevant. The planner cannot bypass the guard; the response slot in
the task graph is cleared and the loop continues until the condition
holds or the soft deadline fires.

\paragraph{Soft deadline.}
A wall-clock soft deadline $T_\textup{max}$ (default $70$ minutes;
configurable via \emph{SOFT\_DEADLINE\_MINUTES}) caps long-running
queries. When it fires, two things happen. \emph{First},
\emph{add\_task()} rejects any new search dispatch, returning the
message ``soft deadline reached: call \emph{finish()} to write the
report''. \emph{Second}, the manager invokes
\emph{\_enforce\_soft\_deadline}, which removes every remaining
\emph{unknown} node from $G_t$ so that the next \emph{finish()} call
passes the convergence guard. The writing pipeline
(\S\ref{sec:writing}) is then invoked on the resulting partial graph.
An independent iteration cap \emph{max\_turn} (default $20$ planning
turns, with a hard ceiling at $3 \!\times\! \textit{max\_turn}$ total
iterations) protects against runaway loops in pathological cases that
escape the wall-clock check.

\paragraph{Action exclusivity.}
The three planner actions \emph{add\_task()},
\emph{propose\_restructure()}, and \emph{finish()} are mutually
exclusive within a single planner turn: each must be issued in its
own code block, with no other actions in the same block. The
\emph{add\_task()} vs.\ \emph{propose\_restructure()} dichotomy
implements the search/restructuring split of Eq.~\ref{eq:dynamics};
\emph{finish()} sits on top of both as a third exclusive option,
applied only at the end of a turn and accepted only when the
convergence guard passes.

\section{Ablation Experiment Details}
\label{app:ablation}
\label{app:ablation-impl}

We report four ablations targeting the three regulatory loops of
\S\ref{sec:intro} plus the topological substrate that schema revision operates on,
together with their full combination $A_\text{full}$. All ablations preserve
the search interface, the Reader's criterion-aligned extraction (Part A), the
evidence store $\mathcal{E}$, and the three-layer writing pipeline
(\S\ref{sec:writing}); modifications are confined to the cognitive layer
defined in \S\ref{sec:cg}--\S\ref{sec:dynamics}. All variants run on the
same 100 DeepResearch Bench queries with Config-27B and identical external
dependencies (Serper, Jina Reader, identical sampling temperature), so that
ablation differences can be attributed to the cognitive layer rather than the system
substrate.

Table~\ref{tab:ablation_components} delineates the precise boundary of each
ablation, allowing reviewers to see at a glance which components are removed
and which are preserved. Each subsection then provides the conceptual
definition together with the rationale for any concomitant modifications.

\begin{table*}[t]
\centering
\small
\setlength{\tabcolsep}{10pt}
\renewcommand{\arraystretch}{1.20}
\begin{tabular}{@{}p{0.55\textwidth} ccccc@{}}
\toprule
\rowcolor{ablHeaderBg}
\textbf{Component} & $A_1$ & $A_2$ & $A_3$ & $A_4$ & $A_\text{full}$ \\
\midrule
\rowcolor{ablGroupBg}
\multicolumn{6}{@{}l}{\textit{Cognitive layer}}\\
Deviation signal $\delta_j^t$ (CR/AAP/$\phi$/$\psi$)
  & \textcolor{rmvdC}{$\times$} & \textcolor{keptC}{$\checkmark$} & \textcolor{keptC}{$\checkmark$} & \textcolor{keptC}{$\checkmark$} & \textcolor{rmvdC}{$\times$} \\
Strategy router $\pi:\mathcal{D}\to\mathcal{A}_\text{search}$
  & \textcolor{rmvdC}{$\times$} & \textcolor{keptC}{$\checkmark$} & \textcolor{keptC}{$\checkmark$} & \textcolor{keptC}{$\checkmark$} & \textcolor{rmvdC}{$\times$} \\
State fusion $\oplus_G$ (Manager LLM reinterpretation)
  & \textcolor{keptC}{$\checkmark$} & \textcolor{keptC}{$\checkmark$} & \textcolor{rmvdC}{$\times$} & \textcolor{keptC}{$\checkmark$} & \textcolor{rmvdC}{$\times$} \\
\textsc{known} state transition
  & \textcolor{keptC}{$\checkmark$} & \textcolor{keptC}{$\checkmark$} & \textcolor{rmvdC}{$\times$} & \textcolor{keptC}{$\checkmark$} & \textcolor{rmvdC}{$\times$} \\
Contradiction tuples / cross-node routing $\mathcal{T}_k^{t+1}$
  & \textcolor{keptC}{$\checkmark$} & \textcolor{keptC}{$\checkmark$} & \textcolor{rmvdC}{$\times$} & \textcolor{keptC}{$\checkmark$} & \textcolor{rmvdC}{$\times$} \\
Restructure ops $\{R_\text{conc}, R_\text{pivot}, R_\text{correct}\}$
  & \textcolor{keptC}{$\checkmark$} & \textcolor{rmvdC}{$\times$} & \textcolor{keptC}{$\checkmark$} & \textcolor{rmvdC}{$\times$}$^\dagger$ & \textcolor{rmvdC}{$\times$} \\
Restructure ops $\{R_\text{aug}, R_\text{prune}\}$
  & \textcolor{keptC}{$\checkmark$} & \textcolor{keptC}{$\checkmark$} & \textcolor{keptC}{$\checkmark$} & \textcolor{keptC}{$\checkmark$} & \textcolor{keptC}{$\checkmark$} \\
Edge set $E_t$ exposed to planner
  & \textcolor{keptC}{$\checkmark$} & \textcolor{keptC}{$\checkmark$} & \textcolor{keptC}{$\checkmark$} & \textcolor{rmvdC}{$\times$} & \textcolor{rmvdC}{$\times$} \\
\texttt{hop\_distance} / \texttt{discovery\_dependency}
  & \textcolor{keptC}{$\checkmark$} & \textcolor{keptC}{$\checkmark$} & \textcolor{keptC}{$\checkmark$} & \textcolor{rmvdC}{$\times$} & \textcolor{rmvdC}{$\times$} \\
Upstream-neighbour context to searcher
  & \textcolor{keptC}{$\checkmark$} & \textcolor{keptC}{$\checkmark$} & \textcolor{keptC}{$\checkmark$} & \textcolor{rmvdC}{$\times$} & \textcolor{rmvdC}{$\times$} \\
\midrule
\rowcolor{ablGroupBg}
\multicolumn{6}{@{}l}{\textit{Shared infrastructure (preserved across all ablations)}}\\
Searcher's structured output with $[\![m]\!]$ citations
  & \textcolor{keptC}{$\checkmark$} & \textcolor{keptC}{$\checkmark$} & \textcolor{keptC}{$\checkmark$} & \textcolor{keptC}{$\checkmark$} & \textcolor{keptC}{$\checkmark$} \\
Reader Part A with verbatim-quote constraint
  & \textcolor{keptC}{$\checkmark$} & \textcolor{keptC}{$\checkmark$} & \textcolor{keptC}{$\checkmark$} & \textcolor{keptC}{$\checkmark$} & \textcolor{keptC}{$\checkmark$} \\
Evidence store $\mathcal{E}$ and $[\![m]\!]$ traceability
  & \textcolor{keptC}{$\checkmark$} & \textcolor{keptC}{$\checkmark$} & \textcolor{keptC}{$\checkmark$} & \textcolor{keptC}{$\checkmark$} & \textcolor{keptC}{$\checkmark$} \\
Three-layer writing pipeline
  & \textcolor{keptC}{$\checkmark$} & \textcolor{keptC}{$\checkmark$} & \textcolor{keptC}{$\checkmark$} & \textcolor{keptC}{$\checkmark$} & \textcolor{keptC}{$\checkmark$} \\
Invariants $\mathbf{I1}$ / $\mathbf{I2}$
  & \textcolor{keptC}{$\checkmark$} & \textcolor{keptC}{$\checkmark$} & \textcolor{keptC}{$\checkmark$} & \textcolor{keptC}{$\checkmark$} & \textcolor{keptC}{$\checkmark$} \\
\bottomrule
\end{tabular}
\caption{Component-level boundary of each ablation.
\textcolor{keptC}{$\checkmark$} preserved; \textcolor{rmvdC}{$\times$} removed.
$^\dagger$: $A_4$ restricts $\mathcal{A}_\text{struct}$ as a logical
consequence of substrate flattening rather than an additional ablation
(see \S\ref{sec:a4_appendix}).}
\label{tab:ablation_components}
\end{table*}

\subsection{$A_1$: Deviation feedback}
\label{sec:a1_appendix}

$A_1$ removes the deviation signal $\delta_j^t$ (Definition~\ref{def:deviation})
and the strategy router $\pi$ (Eq.~\ref{eq:strategy}), forcing the planner to
decide search directions from implicit reasoning over findings alone.
Concretely, the Reader no longer emits CR-AAP scores or the Part-B
\texttt{unexpected\_insights}, the node-level quality profile $\bm{q}_j^t$ is no
longer aggregated, and the five-strategy table (\textsc{exploit} /
\textsc{verify} / \textsc{pivot} / \textsc{explore} / \textsc{substitute}) is
removed from the planner prompt. The prompt-level removal is a necessary
corollary: the strategy router $\pi$ takes $\delta$ as input, and retaining
the strategy table without the deviation signal would leave a dangling
instruction that could spuriously activate strategies on absent input,
contaminating the ablation's semantics. All other mechanisms are preserved:
$\oplus_G$ still interprets findings against the current $G_t$,
$\mathcal{A}_\text{struct}$ remains complete, and the topology remains a graph.

\subsection{$A_2$: Schema revision}
\label{sec:a2_appendix}

$A_2$ restricts $\mathcal{A}_\text{struct}$ from
$\{R_\text{conc}, R_\text{aug}, R_\text{pivot}, R_\text{prune}, R_\text{correct}\}$
to $\{R_\text{aug}, R_\text{prune}\}$, the subset that existing systems can
already realise through outline expansion and section pruning.
$R_\text{conc}$ (concretisation), $R_\text{pivot}$ (axis rotation), and
$R_\text{correct}$ (premise correction) are blocked at three levels (prompt
documentation, API call, and runtime execution), so that even an attempted
invocation by the planner is intercepted. The two-phase safety protocol
(Appendix~\ref{app:safety}) and invariants $\mathbf{I1}$/$\mathbf{I2}$ remain
unchanged, and the rest of the cognitive graph ($\delta$, $\oplus_G$,
topology) is unaffected. Thus $A_2$ isolates the contribution of the three
systematic restructuring operations beyond simple addition and removal.

\subsection{$A_3$: Interpretive update}
\label{sec:a3_appendix}

$A_3$ replaces the interpretive-update operator $\oplus_G$ with text
concatenation. One clarification is needed up front: \textbf{this does not
mean node content becomes unstructured prose}. The Searcher's output is
itself shaped by its own prompt constraints into a coherent inquiry-centred
answer that addresses each acceptance criterion in turn and carries [[$m$]]
citation markers (see the Searcher block in Appendix~\ref{app:prompts}). What $A_3$ removes is the Manager LLM's \emph{secondary
reinterpretation} on top of this output. In the regular update path,
$\oplus_G$ re-examines each finding in the context of $G_t$, classifying it
against $b_{ij}^t$ and $f_j^t$ as criterion-satisfying, redundant,
contradictory, or unexpected; recording contradictions as five-tuples; and
routing cross-node material via $\mathcal{T}_k^{t+1}$. $A_3$ strips this
prior-guided reinterpretation entirely, appending the Searcher's
already-organised output to the target node verbatim.

Because the downstream consumers of $\oplus_G$ are removed alongside it, the
cognitive state $\sigma$ no longer has the basis for a transition to
\textsc{known}. Nodes mechanically advance from \textsc{unknown} to
\textsc{partial} on the first non-empty summary, and edge exhaustion becomes
the sole termination signal. This is again a necessary corollary: the
\textsc{known} judgement depends on the residual pending criteria that
$\oplus_G$ maintains, and without the secondary interpretation those residues
cannot be updated; allowing the transition would let the state machine
advance on inconsistent grounds. The graph topology,
$\mathcal{A}_\text{struct}$, and the evidence store $\mathcal{E}$ are
unchanged.

\subsection{$A_4$: Topological substrate}
\label{sec:a4_appendix}

$A_4$ flattens the structural state $S_t = (N_t, E_t)$ from a graph to a
parallel list of dimensions. The edge set $E_t$ is no longer exposed to the
planner, \texttt{hop\_distance} and \texttt{discovery\_dependency} are
eliminated, and the Searcher receives no upstream-neighbour context. All
LLM-visible prompts are stripped of graph, edge, upstream, downstream, and
hop terminology and reframed in terms of ``research dimensions'', to avoid
leaving conceptual residues that would mislead the planner after the
substrate has changed.

Consistent with the substrate flattening, $A_4$ also restricts
$\mathcal{A}_\text{struct}$ to $\{R_\text{aug}, R_\text{prune}\}$. This
concomitant change is not an additional ablation but a logical consequence of
the substrate change: $R_\text{conc}$ subdivides a node into children with
new edges, $R_\text{pivot}$ rotates dependency direction between nodes, and
$R_\text{correct}$ propagates revisions along downstream subgraphs. All three
are defined over graph structure and have neither conceptual grounding nor an
executable target on a flat list. The Manager LLM and the deviation signal
$\delta$ are unaffected: $\Gamma$ still produces structured updates and
CR-AAP still drives strategy selection. Thus $A_4$ isolates the contribution
of graph topology as a representational substrate.

\subsection{$A_\text{full}$: All four mechanisms ablated}
\label{sec:afull_appendix}

$A_\text{full}$ stacks $A_1$, $A_2$, $A_3$, and $A_4$ simultaneously. The
result is a minimal system that dispatches parallel searches over a flat
list, accumulates the Searcher's output without secondary reinterpretation,
decides next actions from the planner's implicit reading of accumulated text,
and admits only $R_\text{aug}/R_\text{prune}$ when restructuring. The three
regulatory loops and the graph-based cognitive layer are removed, leaving the
search, reading, and writing infrastructure together with basic augmentation
and pruning.

A subtle interaction between two failure modes is worth noting in
$A_\text{full}$. When run alone, $A_3$ contracts the external search ratio to
$0.41\times$ baseline because the Manager no longer tracks pending criteria,
leading the planner to underestimate uncovered gaps and terminate
prematurely. When run alone, $A_1$ inflates the external search ratio to
$1.22\times$ because the absence of deviation signals traps the planner in
inefficient retries. When stacked, $A_3$'s premature termination preempts
$A_1$'s deadlock tendency, yielding an intermediate ratio of $0.54\times$ and
a $0\%$ timeout rate. This complementary-failure pattern is discussed further
in \S\ref{sec:results}.

\section{Per-Dimension Ablation Effects}
\label{app:ablation-perdim}

Table~\ref{tab:ablation-perdim} decomposes each ablation's matched per-query
RACE change against the three-run full-system mean by dimension. The full
ablation affects Insight most strongly ($-4.35$~pp), whereas its effect on
Readability is substantially smaller ($-1.51$~pp).

\begin{table}[t]
\centering
\fontsize{7}{8}\selectfont
\setlength{\tabcolsep}{3pt}
\renewcommand{\arraystretch}{0.85}
\begin{tabularx}{\columnwidth}{l *{4}{>{\centering\arraybackslash}X}}
\toprule
\rowcolor{groupbg}
\textbf{Ablation} & \textbf{Comp.} & \textbf{Insight} & \textbf{Instr.} &
\textbf{Read.} \\
\midrule
A1 & $-1.94$ & $-3.09$ & $-0.61$ & $-1.62$ \\
A2 & $-1.55$ & $-1.37$ & $-1.36$ & $-1.26$ \\
A3 & $-2.28$ & $-2.08$ & $-1.20$ & $-1.51$ \\
A4 & $-1.44$ & $-1.09$ & $-0.70$ & $-1.70$ \\
A\textsubscript{full} & $-2.89$ & $-4.35$ & $-1.36$ & $-1.51$ \\
\bottomrule
\end{tabularx}
\caption{Matched per-query RACE changes in percentage points against the
three-run full-system mean.}
\label{tab:ablation-perdim}
\end{table}

\section{Ablation Effects Across Backbone Scale}
\label{app:scale-ablation}

We rerun A2 and A4 under Config-Plus on the 26 queries that triggered
$R_\textup{conc}$, $R_\textup{pivot}$, or $R_\textup{correct}$ in at least one
of the three Config-27B runs. We use the same propensity groups and per-query
$\Delta$ calculation as in \autoref{tab:ablation-accomm}. Config-27B uses its
three-run full-system mean, whereas Config-Plus uses its full-system run as the
baseline.

\begin{table*}[t]
\centering
\fontsize{8}{9}\selectfont
\setlength{\tabcolsep}{6pt}
\renewcommand{\arraystretch}{0.9}
\begin{tabularx}{\textwidth}{l *{4}{>{\centering\arraybackslash}X}}
\toprule
\rowcolor{groupbg}
\textbf{Ablation} & \textbf{27B $p\geq2/3$} & \textbf{27B $p=1/3$} &
\textbf{Plus $p\geq2/3$} & \textbf{Plus $p=1/3$} \\
\midrule
A2 & $-4.61$ & $-1.52$ & $-1.43$ & $-0.70$ \\
A4 & $-2.98$ & $-0.64$ & $-1.36$ & $-1.36$ \\
\textbf{A4$-$A2 gap ($p\geq2/3$)} & \textbf{$+1.63$} & &
\textbf{$+0.08$} & \\
\bottomrule
\end{tabularx}
\caption{Schema-revision and topology ablations across backbone scale and
restructure propensity. Each value is the matched per-query RACE Overall change
(pp) from the corresponding backbone's full system. The gap is calculated from
the unrounded, separately estimated A4 and A2 group means.}
\label{tab:scale-ablation}
\end{table*}

\section{Blind Human Evaluation}
\label{app:human_eval}

Four non-author PhD annotators independently compare the first-run outputs of VeriTrace and WebWeaver on
20 DeepResearch Bench queries (12 Chinese and 8 English). Both systems use the
same Qwen3.5-27B writer. Reports are randomised as A/B for each query and
de-anonymised only after scoring. Annotators record win, tie, or loss for the
four RACE dimensions. Table~\ref{tab:human_eval} reports the annotator-level
win/tie/loss counts and net win rates.

\begin{strip}
\centering
\fontsize{8}{9}\selectfont
\setlength{\tabcolsep}{6pt}
\renewcommand{\arraystretch}{0.9}
\begin{tabularx}{\textwidth}{l *{5}{>{\centering\arraybackslash}X}}
\toprule
\rowcolor{groupbg}
\textbf{Annotator} & \textbf{Comp.} & \textbf{Insight} & \textbf{Instr.} &
\textbf{Read.} & \textbf{Net win} \\
\midrule
1 & 4/16/0 & 12/8/0 & 6/13/1 & 7/13/0 & $+35\%$ \\
2 & 11/6/3 & 17/3/0 & 8/10/2 & 14/6/0 & $+56\%$ \\
3 & 6/8/6 & 13/6/1 & 12/4/4 & 17/3/0 & $+46\%$ \\
4 & 6/11/3 & 13/7/0 & 6/10/4 & 8/10/2 & $+30\%$ \\
\bottomrule
\end{tabularx}
\captionof{table}{Blind human comparison of VeriTrace against WebWeaver. Each dimension
reports win/tie/loss counts from the VeriTrace perspective over 20 queries.}
\label{tab:human_eval}
\end{strip}

\section{Case Study: Sovereign Wealth Investment (Query 53)}
\label{app:casestudy}

We trace one full assimilation/accommodation cycle on DeepResearch Bench query~53 (``Researching how the world's wealthiest governments invest'') to make the three loops concrete on a single trajectory. The query is illustrative because its first-round search exposes a textbook \emph{concept-vs-entity} mismatch in the initial framing: the planner had assumed a single ``wealthiest governments'' node feeding two abstract dimension nodes, but the evidence actually co-locates by investing entity. The system recognises the mismatch, proposes $R_\textup{conc}$, and after restructuring continues on a topology that compresses what would have been $4 \times 2 = 8$ task slots into 4 entity-focused investigations. Final RACE Overall is $0.578$ (Insight $0.620$), placing the query in the top quartile of VeriTrace's English-language runs.

\subsection{Trajectory at a glance}

The full trajectory comprises 9 planner iterations across 4 search turns, summarised in Table~\ref{tab:case_q53_trajectory}. The following subsections cover the $R_\textup{conc}$ decision in \S\ref{ssec:q53_rconc}, deviation-driven strategy routing in \S\ref{ssec:q53_strategy}, and the termination guard and three-layer writing pipeline in \S\ref{ssec:q53_writing}.

\begin{table*}[t!]
\centering
\scriptsize
\setlength{\tabcolsep}{3pt}
\renewcommand{\arraystretch}{1.2}
\begin{tabular}{@{}c >{\raggedright\arraybackslash}p{0.28\textwidth} >{\raggedright\arraybackslash}p{0.60\textwidth}@{}}
\toprule
\rowcolor{groupbg}\textbf{Turn (iter)} & \textbf{Planner action} & \textbf{Outcome / signal} \\
\midrule
0     & parse $G_0$ & 4 nodes, 3 edges; $e_2,e_3,e_4$ \textsc{unknown} \\
1     & \texttt{add\_task} on $r_1$                                                                         & 42 URLs, 13 findings on $e_2$;\ \  $\overline{\textup{cr}}\!=\!3.6$, $\overline{\textup{aap}}\!=\!3.7$, find\,=\,strong \\
2~(r) & reflect, no code (Figure~\ref{fig:case_q53_planner_reflection})                                       & co-occurrence diagnosis on $e_2$; candidate ops $R_\textup{conc}$\,+\,$R_\textup{pivot}$ resolved to single $R_\textup{conc}$ \\
2~(a) & \texttt{propose\_restructure}($R_\textup{conc}$) (Figure~\ref{fig:case_q53_restructure_log})          & Phase~1: +4 nodes, +4 edges; cascade-remove $e_3,e_4$; violations\,=\,0 \\
3     & 4 parallel \texttt{add\_task}s on $r_\textup{dyn,1..4}$                                             & all 4 nodes $\to$ \textsc{partial}; $\pi\!=\!$\textsc{exploit} on each \\
4--6  & 4 follow-ups (3 \textsc{exploit}, 1 \textsc{pivot})                                                 & $e_\textup{dyn,k}\!\to\!\textsc{known}$; \texttt{task\_gov\_fiscal\_health\_ranking} writes back to $e_2$ via $\mathcal{T}_2^{t+1}$ \\
7     & termination reflection (Figure~\ref{fig:case_q53_termination}) $\to$ \texttt{finish()}                & $G_T$: 6 nodes, $594$ evidence records, $103$ unique $[\![m]\!]$ \\
8     & writing pipeline (3 layers)                                                                         & 8 sections, $52$ insights, $25$K-char report \\
\bottomrule
\end{tabular}
\caption{Query~53 trajectory. Each row is one planner iteration; Turn~2 is split into reflect (\textsc{r}) and act (\textsc{a}) sub-iterations to show the two-step pattern. Figure cross-references in the planner-action column point to the verbatim reasoning logs shown in this appendix. Source: run log \texttt{q053\_parallel\_*.log}.}
\label{tab:case_q53_trajectory}
\end{table*}

\subsection{Initial graph and the $R_\textup{conc}$ trigger}\label{ssec:q53_rconc}

The Cognitive Graph Manager parses the query into four entities and three edges (Figure~\ref{fig:case_q53_before}, left panel). $e_1$ (\emph{start}) anchors the user question; $e_2$ is a placeholder for the wealthiest-governments list; $e_3$ asks how those governments \emph{operate} their wealth; $e_4$ asks how they \emph{allocate} assets. Edges $r_2\!:\!e_2\!\to\!e_3$ and $r_3\!:\!e_2\!\to\!e_4$ encode the planner's prior assumption that two parallel dimensional investigations follow once $e_2$ is populated.

The planner's first \texttt{add\_task} call on $r_1$ targets $e_2$ with two core acceptance criteria; after the search executes and the Cognitive Graph Manager assimilates the response, $e_2$ carries 13 \texttt{discovered\_items} indexed by global $[\![m]\!]$ refs, plus two Reader Part-B \texttt{unexpected\_insights}. Figure~\ref{fig:case_q53_round1} reproduces both the issued task and the resulting compiled graph state that the planner reads on the next turn.

\begin{figure*}[!htbp]
\begin{gmemorycasebox}{Round 1: first task and resulting graph state}
\begin{gmemcontent}

\noindent\textbf{(a) Planner's first \texttt{add\_task} on $r_1$:}
\begin{redlog}
add_task(
    node_name = "task_wealthiest_governments_definition_ranking",
    cognitive_edge_id = "r1",  cognitive_target_id = "e2",
    task_type = "open",
    core_criteria = [
        "definition and indicators of governmental wealth",
        "country/government wealth rankings and top 10"],
    supplementary_criteria = [
        "data-source institutions",
        "consistency across indicators",
        "composite assessment methods"],
    node_content = """
    Goal: explore the definition of 'world's wealthiest
    governments' and the currently accepted ranking.
    Constraints: prefer 2023-2024 data; international
    organisations / academic / authoritative finance
    media. Do NOT presuppose a single canonical
    definition -- let the search reveal how the concept
    is operationalised in practice.""")
\end{redlog}

\medskip
\noindent\textbf{(b) Cognitive graph state after Round 1} (compiled view, what the Planner reads next turn):
\begin{redlog}
## Current Cognitive Graph State
**Structure type**: convergence

### Entity Nodes
- [START]   User question                  (e1, hop=0)
- [PARTIAL] Wealthiest governments list    (e2, hop=1)
  - Discovered items (top-GDP):
    US [[1]], China [[1]], Germany [[2]], Japan [[2]],
    UK [[7]], India [[6]], France [[2]], ...
  - Discovered items (SWF-managers):
    PBoC ($3.7T) [[16]], NBIM ($2.1T) [[17]],
    SAFE ($1.95T) [[16]], GPIF ($1.88T) [[19]],
    CIC (\$1.57T) [[17]]
  - Cross-item insights:
    - definitional split: GDP vs gov assets [[6]][[33]]
    - top-3 SWFs control >50% of global SWF AUM [[18]]
  - Unexpected (Reader Part-B):
    - state pensions may exceed nat'l SWFs [[35]]
    - shell companies distort GDP figures [[3]]
  - Quality: CR=3.6 AAP=3.7  finding=strong, unexpected=moderate
  - Supplementary pending: gov_balance_sheet_rankings

### To-solve inquiry goals
- e2 --operating_mechanism--> e3   (r2, 0/3)
- e2 --asset_allocation-->    e4   (r3, 0/3)
\end{redlog}

\end{gmemcontent}
\end{gmemorycasebox}
\caption{What the Planner sees at the start of Turn~2. (a) The first \texttt{add\_task} dispatched on edge $r_1$ targets $e_2$ with two core acceptance criteria. (b) After the search executes and the Cognitive Graph Manager assimilates, $e_2$ carries 13 \texttt{discovered\_items} indexed by global $[\![m]\!]$ refs, plus two Reader Part-B \texttt{unexpected\_insights}. This is the input on which the planner emits the reflection in Figure~\ref{fig:case_q53_planner_reflection}.}
\label{fig:case_q53_round1}
\end{figure*}

After Round~1 on $r_1$, the planner detects a structural rather than informational gap. The deviation profile shows $e_2$ accumulating 13 strong findings (GDP-leading top-10, SWF top-5, plus Reader Part-B \texttt{unexpected\_insights} on state-level pensions exceeding national SWFs and foreign shell companies distorting GDP figures), while downstream dimensions would inherit none of the per-entity distribution they require; Reader pages consistently bundle ``operating mechanism + asset allocation'' by investing entity rather than by dimension. Rather than searching, the planner emits its full deviation analysis as a thinking-only iteration with no executable code (Figure~\ref{fig:case_q53_planner_reflection}), identifying two candidate operators, $R_\textup{conc}$ (concretise $e_2$) and $R_\textup{pivot}$ (rotate the dimension-vs-entity axis), and resolving that $R_\textup{conc}$ with bundled \texttt{core\_criteria} already subsumes $R_\textup{pivot}$, so a single operator suffices.

\begin{figure*}[!p]
\begin{gmemorycasebox}{Planner reflection on round-1 findings (verbatim from log, translated)}
\begin{gmemcontent}
\textbf{Findings review}:
\begin{redlog}
e2 now contains substantive findings:
- definitional split: GDP vs gov assets
- GDP top-10 (US, CN, DE, JP, UK, ...)
- SWF top-5 (PBoC, NBIM, SAFE, GPIF, CIC)
- unexpected_insights (Reader Part-B):
  * state pensions may exceed nat'l SWFs
  * shell companies distort GDP figures
\end{redlog}

\textbf{Downstream impact and concretisation}:
\begin{redlog}
- r2 (op-mechanism): per-entity inquiry,
  info lives in different sources.
- r3 (asset-allocation): per-entity too.

Conclusion: e2 should be concretised.
Two classes:
  (1) GDP-leading economies (~10)
  (2) SWF managers (~5-6, e.g. China)
=> integrate to ~8-10 investing subjects.
\end{redlog}

\textbf{Axis-rotation diagnosis}:
\begin{redlog}
r2/r3 info co-occurs BY SUBJECT,
not by dimension. Reports on NBIM
cover both operation AND allocation;
no single source covers all subjects
on one dimension.
=> fold r2/r3 criteria into each
   subject's inquiry, avoid N x M.
\end{redlog}

\textbf{Restructure decision}:
\begin{redlog}
op_type = R_conc + R_pivot
=> R_conc with bundled core_criteria
   absorbs the rotation; one operator
   suffices for this state.
\end{redlog}
\end{gmemcontent}
\end{gmemorycasebox}
\caption{The planner's free-form deviation analysis on the round-1 findings, emitted as a thinking-only iteration with no executable code. Reader Part-B \texttt{unexpected\_insights} feed into the diagnosis alongside Part-A acceptance criteria. The planner identifies two candidate operators ($R_\textup{conc}$ and $R_\textup{pivot}$) and resolves that one absorbs the other under bundled \texttt{core\_criteria}. The structured \texttt{propose\_restructure} call follows on the next iteration (Figure~\ref{fig:case_q53_restructure_log}).}
\label{fig:case_q53_planner_reflection}
\end{figure*}

On the next iteration the planner commits this analysis to a structured \texttt{propose\_restructure} call. The Manager's evaluation independently echoes the same dual-operator framing (``concretisation plus axis rotation''), and its Phase~1 edits are reproduced verbatim from the log. Figure~\ref{fig:case_q53_restructure_log} also shows the four parallel \texttt{add\_task} calls dispatched immediately after the restructure, so the reader can see what the new topology turns into operationally.

\begin{figure*}[!htbp]
\begin{gmemorycaseboxfixed}{R$_{\textup{conc}}$ execution on $e_2$ and Turn 3 dispatch (verbatim from log)}
\begin{gmemcontent}
\textbf{Planner intent} (\texttt{op\_type=R\_conc}):
\begin{redlog}
"world's wealthiest governments" actually appears as two
classes of subject: (1) top-GDP economies and (2) sovereign
wealth managers (Norway NBIM \$2.1T, China PBoC/SAFE/CIC
\$7.2T, Japan GPIF \$1.9T). The two downstream dimensions
co-occur by subject, not by dimension. Concretise e2 into
4 subject nodes; fold r2/r3 criteria into each subject's
inquiry, collapsing 4x2=8 prospective tasks into 4.
\end{redlog}

\textbf{Manager Phase 1 edits} (\texttt{added=4, removed=2, violations=0}):
\begin{redlog}
+ add_node  e_dyn_1  Norway NBIM (\$2.1T)
+ add_node  e_dyn_2  China SWF system (PBoC/SAFE/CIC, \$7.2T)
+ add_node  e_dyn_3  Japan GPIF (\$1.9T)
+ add_node  e_dyn_4  US Federal investment system
+ add_edge  r_dyn_1 .. r_dyn_4   (e2 -> e_dyn_k)
- remove_node  e3  (cascade: r2)
- remove_node  e4  (cascade: r3)

protected = {e1, e2}      (N_user)
preserved = f_2^t        (I1: 13 findings on e2 untouched)
\end{redlog}

\textbf{Manager evaluation:}
\begin{redlog}
"Search has identified four primary investing subjects;
downstream dimensions co-occur by subject. Concretisation
plus axis rotation into 4 focused tasks is a reasonable
cognitive-graph evolution."
\end{redlog}

\medskip
\noindent\textbf{Aftermath: Turn 3 dispatches 4 parallel tasks (one per new entity):}
\begin{redlog}
add_task(node_name = "task_nbim_investment_profile",
    cognitive_target_id = "e_dyn_1",  task_type = "open",
    core_criteria = [
        "fund size and AUM trajectory",
        "operating mechanism (mandate, governance)",
        "asset allocation by class and geography"],
    ...)

add_task(node_name = "task_china_swf_investment_profile",
    cognitive_target_id = "e_dyn_2",  task_type = "open",
    core_criteria = [
        "PBoC/SAFE/CIC division of labour",
        "operating mechanism per entity",
        "asset allocation per entity"],
    ...)

# (+ task_gpif_investment_profile, task_us_federal_investment_profile)
\end{redlog}
\end{gmemcontent}
\end{gmemorycaseboxfixed}
\caption{The complete $R_\textup{conc}$ trace from the run log, plus the four parallel \texttt{add\_task}s the planner dispatches on the next turn. Acceptance criteria of the deleted dimension nodes are folded into each new edge's \texttt{core\_criteria}, so each $e_\textup{dyn,k}$ inherits both ``operating mechanism'' and ``asset allocation'' as bundled criteria. Phase~2 reachability check passes with no orphans.}
\label{fig:case_q53_restructure_log}
\end{figure*}

The before/after structural change is visualised in Figure~\ref{fig:case_q53_before}: $e_3$ and $e_4$ are cascade-removed, four entity-focused nodes $e_\textup{dyn,1\dots4}$ branch from $e_2$, and the bundled criteria of the deleted edges are inherited by each new edge.

\begin{figure*}[h]
\centering
\begin{tikzpicture}[
  font=\footnotesize,
  entity/.style={rectangle, rounded corners=2pt, draw=ourblue, thick,
    fill=ourbg, minimum width=20mm, minimum height=8mm, align=center, inner sep=2pt},
  start/.style={rectangle, rounded corners=2pt, draw=ourblue, very thick,
    fill=ourblue, text=white, minimum width=16mm, minimum height=7mm, align=center,
    font=\footnotesize\bfseries, inner sep=2pt},
  preserved/.style={rectangle, rounded corners=2pt, draw=ourblue, thick,
    fill=fsbg, minimum width=22mm, minimum height=8mm, align=center, inner sep=2pt},
  newentity/.style={rectangle, rounded corners=2pt, draw=ourblue, thick,
    fill=ourbg, minimum width=20mm, minimum height=10mm, align=center, inner sep=2pt},
  removed/.style={rectangle, rounded corners=2pt, draw=gray!70, dashed, thick,
    fill=white, minimum width=18mm, minimum height=7mm, align=center,
    text=gray!70, inner sep=2pt},
  edge/.style={->, >=Stealth, semithick, draw=ourblue!80!black},
  newedge/.style={->, >=Stealth, semithick, draw=ourblue},
  panellabel/.style={font=\scriptsize\bfseries, color=ourblue}
]
\begin{scope}[local bounding box=panelL]
\node[panellabel] at (0, 1.5) {(a)~$G_0$ : initial graph (4 nodes, 3 edges)};
\node[start] (e1) at (0, 0.6) {$e_1$ : START};
\node[entity] (e2) at (0, -1.0) {$e_2$ : wealthiest\\governments\\\scriptsize\itshape (placeholder)};
\node[entity] (e3) at (-2.2, -2.9) {$e_3$ : op-\\mechanism};
\node[entity] (e4) at (2.2, -2.9) {$e_4$ : asset\\allocation};
\draw[edge] (e1) -- node[right=1pt, font=\scriptsize] {$r_1$} (e2);
\draw[edge] (e2) -- node[above left=-1pt, font=\scriptsize] {$r_2$} (e3);
\draw[edge] (e2) -- node[above right=-1pt, font=\scriptsize] {$r_3$} (e4);
\node[font=\scriptsize\itshape, text=gray!70, align=center] at (0, -3.9)
  {$4 \times 2 = 8$ prospective task slots};
\end{scope}
\node[font=\large\bfseries, color=ourblue] at (4.0, -1.1) {$\Longrightarrow$};
\node[font=\scriptsize, color=ourblue, align=center] at (4.0, -1.9)
  {$R_\textup{conc}$};
\begin{scope}[xshift=7.8cm, local bounding box=panelR]
\node[panellabel] at (0, 1.5) {(b)~$G_t$ after $R_\textup{conc}$ : 4 entity-focused inquiries};
\node[start] (e1r) at (0, 0.6) {$e_1$ : START};
\node[preserved] (e2r) at (0, -1.0)
  {$e_2$ (\textbf{I1}: $f_2$ preserved)\\\scriptsize\itshape 13 findings retained};
\node[removed] (e3r) at (-3.0, -2.5) {$e_3$ : op-mech};
\node[removed] (e4r) at (3.0, -2.5) {$e_4$ : alloc};
\draw[gray!70, thick] (e3r.south west) -- (e3r.north east);
\draw[gray!70, thick] (e3r.north west) -- (e3r.south east);
\draw[gray!70, thick] (e4r.south west) -- (e4r.north east);
\draw[gray!70, thick] (e4r.north west) -- (e4r.south east);
\node[newentity] (d1) at (-3.6, -4.0) {$e_\textup{dyn,1}$\\Norway\\NBIM};
\node[newentity] (d2) at (-1.2, -4.0) {$e_\textup{dyn,2}$\\China SWF};
\node[newentity] (d3) at (1.2, -4.0) {$e_\textup{dyn,3}$\\Japan\\GPIF};
\node[newentity] (d4) at (3.6, -4.0) {$e_\textup{dyn,4}$\\US Federal};
\draw[edge] (e1r) -- (e2r);
\draw[newedge] (e2r) -- (d1);
\draw[newedge] (e2r) -- (d2);
\draw[newedge] (e2r) -- (d3);
\draw[newedge] (e2r) -- (d4);
\node[draw=ourblue, dashed, rounded corners=2pt, fill=ourbg, inner sep=3pt,
      font=\scriptsize, align=center, text=ourblue] at (0, -5.3)
  {each $r_\textup{dyn,k}$ inherits bundled criteria $\{$op-mech, asset-alloc$\}$\\
   $\Rightarrow$ 8 prospective slots collapse to $\boldsymbol{4}$ entity-focused tasks};
\end{scope}
\end{tikzpicture}
\caption{Cognitive graph before (a) and after (b) $R_\textup{conc}$ on query~53. Findings on $e_2$ are preserved (\textbf{I1}); $e_3, e_4$ are cascade-removed because their inquiry criteria fold into the new entity-focused edges, turning $4 \times 2 = 8$ prospective task slots into $4$.}
\label{fig:case_q53_before}
\end{figure*}

\subsection{Deviation signal, strategy router, next task}\label{ssec:q53_strategy}

After restructuring, the four entity-focused tasks run in parallel. We zoom into \texttt{task\_nbim\_investment\_profile} to make the deviation-to-strategy chain concrete. A representative high-quality page (NBIM's own Government Pension Fund Global annual report) scores $(c, r, \alpha, \beta, \rho) = (4, 4, 5, 4, 3)$, yielding via Eq.~\ref{eq:cr-page}--\ref{eq:aap-page} the page-level composites $\textup{cr}_m = 4.00$ and $\textup{aap}_m = 4.15$. Aggregating across the $13$ pages of the task gives the node profile $(\overline{\textup{cr}}_{\textup{dyn,1}}, \overline{\textup{aap}}_{\textup{dyn,1}}) = (3.9, 4.4)$ with \texttt{finding\_strength=strong} and no accessibility barriers. The strategy router $\pi$ then evaluates the predicates of Eq.~\ref{eq:strategy} in priority order; Figure~\ref{fig:case_q53_craap} traces the full Likert-to-composite-to-decision chain.

\begin{figure*}[h]
\centering
\begin{tikzpicture}[
  font=\footnotesize,
  panellabel/.style={font=\scriptsize\bfseries, color=ourblue},
  bar/.style={fill=ourblue, draw=ourblue!80!black, thick},
  decactive/.style={rectangle, rounded corners=3pt, draw=ourblue, very thick,
    fill=ourbg, minimum width=16mm, minimum height=6.5mm,
    align=center, font=\scriptsize\bfseries, inner sep=2pt, text=ourblue},
  termexploit/.style={rectangle, rounded corners=3pt, draw=red!70!black, very thick,
    fill=red!8, minimum width=18mm, minimum height=7mm,
    align=center, font=\scriptsize\bfseries, inner sep=2pt, text=red!70!black},
  termother/.style={rectangle, rounded corners=3pt, draw=gray!70, dashed,
    fill=white, minimum width=16mm, minimum height=6mm,
    align=center, font=\scriptsize, inner sep=2pt, text=gray!70},
  edgeactive/.style={->, >=Stealth, very thick, draw=red!70!black},
  edgeinactive/.style={->, >=Stealth, semithick, draw=gray!70, dashed}
]
\begin{scope}[local bounding box=panelL]
\node[panellabel] at (2.8, 4.6) {(a)~Page rating $\to$ composite scores};
\foreach \x/\lab in {0/0, 1/1, 2/2, 3/3, 4/4, 5/5} {
  \draw[gray!40, very thin] (\x, 0.2) -- (\x, 3.9);
  \node[font=\scriptsize, text=gray!70] at (\x, 0.0) {\lab};
}
\node[font=\scriptsize, text=gray!70] at (2.5, -0.4) {Likert score};
\foreach \y/\dim/\val/\desc in {%
  3.6/c/4/cred,
  3.0/r/4/rel,
  2.4/$\alpha$/5/acc,
  1.8/$\beta$/4/purp,
  1.2/$\rho$/3/rec} {
  \node[font=\scriptsize, anchor=east] at (-0.15, \y) {\dim};
  \draw[bar] (0, \y-0.18) rectangle (\val, \y+0.18);
  \node[font=\scriptsize, anchor=west, text=ourblue] at (\val+0.1, \y) {\val};
  \node[font=\tiny\itshape, anchor=west, text=gray!70] at (5.2, \y) {\desc};
}
\node[draw=ourblue, thick, rounded corners=2pt, fill=ourbg,
      align=left, inner sep=3pt, anchor=north west, font=\scriptsize,
      text width=58mm] (compbox) at (-0.4, -0.7)
{\textbf{page composites} (Eq.~\ref{eq:cr-page}--\ref{eq:aap-page}):\\
   ~~$\textup{cr}_m = 0.30{\cdot}4 + 0.70{\cdot}4 = \mathbf{4.00}$\\
   ~~$\textup{aap}_m = 0.40{\cdot}5 + 0.35{\cdot}4 + 0.25{\cdot}3 = \mathbf{4.15}$};
\node[draw=ourblue, dashed, rounded corners=2pt, fill=fsbg,
      align=left, inner sep=3pt, anchor=north west, font=\scriptsize,
      text width=58mm,
      below=1.5mm of compbox.south west, anchor=north west]
  {\textbf{node profile} $e_\textup{dyn,1}$ (over 13 pages):\\
   ~~$(\overline{\textup{cr}}, \overline{\textup{aap}}) = (3.9, 4.4)$,~finding\,=\,\textsf{strong},~$\psi=$\textsf{none},~$\phi=0$};
\end{scope}
\begin{scope}[xshift=8.8cm, local bounding box=panelR]
\node[panellabel] at (2.5, 4.6) {(b)~Strategy router $\pi(\delta)$ on $e_\textup{dyn,1}$};
\node[decactive] (root) at (2.5, 3.7) {Accessible?};
\node[decactive] (rel) at (1.0, 2.5) {Relevant?};
\node[termother] (substitute) at (4.2, 2.5) {\textsc{substitute}};
\node[decactive] (cred) at (-0.3, 1.3) {Credible?};
\node[termother] (pivot) at (2.4, 1.3) {\textsc{pivot}};
\node[termexploit] (exploit) at (-1.4, 0.1) {\textsc{exploit}};
\node[termother] (verify) at (1.0, 0.1) {\textsc{verify}};
\draw[edgeactive] (root) -- node[above left=-2pt, font=\scriptsize\bfseries, text=red!70!black] {y} (rel);
\draw[edgeinactive] (root) -- node[above right=-2pt, font=\scriptsize] {n} (substitute);
\draw[edgeactive] (rel) -- node[above left=-2pt, font=\scriptsize\bfseries, text=red!70!black] {y} (cred);
\draw[edgeinactive] (rel) -- node[above right=-2pt, font=\scriptsize] {n} (pivot);
\draw[edgeactive] (cred) -- node[above left=-2pt, font=\scriptsize\bfseries, text=red!70!black] {y} (exploit);
\draw[edgeinactive] (cred) -- node[above right=-2pt, font=\scriptsize] {n} (verify);
\node[draw=red!70!black, thick, rounded corners=2pt, fill=red!8,
      align=center, inner sep=3pt, font=\scriptsize, text=red!70!black] at (2.0, -1.3)
  {next \texttt{add\_task}: \texttt{task\_nbim\_esg\_thresholds}\\
   (\textsc{specified-source}, narrowed criterion)};
\end{scope}
\end{tikzpicture}
\caption{From a Reader page rating to a strategy choice on $e_\textup{dyn,1}$. The path Accessible-yes $\to$ Rel-yes $\to$ Cred-yes selects \textsc{exploit}, which routes the next \texttt{add\_task} to a narrowed sub-criterion within the same source family.}
\label{fig:case_q53_craap}
\end{figure*}

\noindent The strategy router's choice materialises as the following \texttt{add\_task} call:
\begin{redlog}
add_task(node_name = "task_nbim_esg_thresholds",
    cognitive_edge_id = "r_dyn_1",  cognitive_target_id = "e_dyn_1",
    task_type = "specified",
    specified_source = "NBIM official exclusion guidelines",
    core_criteria = ["ESG exclusion quantitative thresholds"],
    supplementary_criteria = ["review process and decision mechanism"],
    node_content = """Find specific quantitative thresholds in
    NBIM ethical screening: coal-revenue %, nuclear/weapons
    revenue red lines, environmental/human-rights triggers
    requiring committee review.""")
\end{redlog}
\noindent The \texttt{task\_type="specified"} field and named source implement \textsc{exploit} by narrowing the search within the same source family.

Across the four new nodes, the procedure yields three \textsc{exploit}-class follow-ups. On $e_2$, where Reader Part-B flagged a missing fiscal-health dimension, the planner instead routes to \textsc{pivot} and creates \texttt{task\_gov\_fiscal\_health\_ranking}, which writes back to $e_2$ via cross-node routing $\mathcal{T}_2^{t+1}$ (Definition~\ref{def:node}). A single $(\overline{\textup{cr}}, \overline{\textup{aap}}, \phi, \psi)$ tuple per node thus yields four different next-task templates across four nodes in the same turn: the deviation signal is an active selector, not a passive monitor.

At convergence, the same selector triggers termination rather than \textsc{exploit} or \textsc{pivot}. After 4 entity-focused tasks plus 4 follow-ups (3 \textsc{exploit}, 1 \textsc{pivot}), all 5 entity nodes reach \textsc{partial} or \textsc{known}, but Searcher feedback continues to flag \texttt{can\_improve=True} on each remaining edge. The planner's termination reflection (Figure~\ref{fig:case_q53_termination}) then partitions the residual gaps into search-recoverable and structurally-unrecoverable categories, and emits \texttt{finish()} on the basis of the latter rather than continuing to expend budget on signals that further search cannot close.

\begin{gmemorycasebox}{Termination reasoning (verbatim log)}
\begin{gmemcontent}
\textbf{Termination check}:
\begin{redlog}
- 0 UNKNOWN nodes; all 5 entity nodes
  (e1, e2, e_dyn_1..4) PARTIAL or KNOWN.
- 4 investigating edges at 2/3 attempts;
  Searcher feedback still flags
  can_improve=True on every edge.
\end{redlog}

\textbf{Capability-boundary diagnosis}:
\begin{redlog}
r1 (gov fiscal-health rankings):
  structural absence: IMF and World Bank
  hold the data but publish no unified
  ranking.

r_dyn_1 (NBIM coal-revenue threshold):
  disclosure form: NBIM exclusion is
  qualitative, not quantitative.

r_dyn_2 (IMF submission technical channel):
  sensitivity: API/FTP specs kept private.

r_dyn_3 (GPIF voting-against trigger):
  access barrier: deep PDF content beyond
  Searcher's page_limit.
\end{redlog}

\textbf{Decision}:
\begin{redlog}
Residual gaps are structural / disclosure
/ sensitivity / access barriers, not
search-strategy failures. Further search
would inflate cost without raising recall.
=> emit finish().
\end{redlog}
\end{gmemcontent}
\end{gmemorycasebox}
\captionof{figure}{Planner reasoning at the convergence guard. The deviation profile distinguishes search-recoverable gaps from gaps that further search cannot close (structural absence, disclosure form, sensitivity, access barriers), allowing the planner to emit \texttt{finish()} rather than continuing to expend budget against \texttt{can\_improve=True} signals.}
\label{fig:case_q53_termination}

\subsection{Three writing layers on a 6-node graph}\label{ssec:q53_writing}

On termination $G_T$ has $6$ nodes, $594$ evidence records in $\mathcal{E}$, and $103$ unique global $[\![m]\!]$ refs. The three-layer pipeline narrows information access in lockstep with decision authority (Figure~\ref{fig:case_q53_writing}). \emph{Layer~1 (OutlinePlanner)} sees the full graph but only finding summaries (no quotes) and the per-node evidence-availability index, and emits an outline of $8$ sections each bound to a node subset $V_k$ (e.g.~$V_7 = \{e_\textup{dyn,1}, e_\textup{dyn,3}, e_2\}$ for the governance-evolution section). \emph{Layer~2 (SectionPlanner)} runs sequentially across §1--§8; each planner sees only its section's evidence index $\{(m, \ell_m)\}$ restricted to $\bigcup_{n_k \in V_k} \mathcal{E}_k$, and returns insights as $(\textit{claim}, \mathcal{I}, \textit{hint})$ tuples with $|\mathcal{I}| \in [2, 5]$ ($52$ insights total on this run). On §7, the planner attempted to bind ref $[\![64]\!]$ to an insight on Norway's exclusion-decision pause; the post-filter caught that $[\![64]\!]$ was outside $\mathcal{E}_7$ and stripped it before the insight reached the writer (logged verbatim as \texttt{[SectionPlanner] Removed invalid evidence refs \{64\}}). \emph{Layer~3 (SectionWriter)} runs sequentially with each writer seeing only the verbatim quotes $v_m$ for its section's bound $\mathcal{I}$.

\begin{figure*}[h]
\centering
\begin{tikzpicture}[
  font=\footnotesize,
  layer/.style={trapezium, draw=ourblue, thick, align=center,
    minimum height=11mm, inner sep=3pt,
    trapezium left angle=72, trapezium right angle=108},
  sidebar/.style={rectangle, draw=gray!70, dashed, rounded corners=2pt,
    fill=groupbg, align=left, inner sep=3pt, font=\scriptsize, text=gray!60!black}
]
\node[layer, fill=ourbg, minimum width=72mm] (L1) at (0, 3.4)
  {\textbf{Layer 1 \textendash\ OutlinePlanner}\\
   \scriptsize sees full graph + finding summaries (no quotes)\\
   \scriptsize emits 8 sections, each bound to $V_k \subseteq N_T$};
\node[layer, fill=ourbg!70!ourblue!20, minimum width=54mm,
      below=2mm of L1] (L2)
  {\textbf{Layer 2 \textendash\ SectionPlanner} (sequential $\times 8$)\\
   \scriptsize sees only $\mathcal{E}_k$ index $\{(m,\ell_m)\}$ (no quotes)\\
   \scriptsize emits 52 insights as $(\textit{claim}, \mathcal{I}, \textit{hint})$, $|\mathcal{I}|\in[2,5]$};
\node[layer, fill=ourblue!30!white, minimum width=36mm,
      below=2mm of L2] (L3)
  {\textbf{Layer 3 \textendash\ SectionWriter} (sequential $\times 8$)\\
   \scriptsize sees verbatim quotes $v_m$ for bound $\mathcal{I}$ only\\
   \scriptsize renders 25K-char report with $[\![m]\!]$ markers};
\draw[->, >=Stealth, very thick, draw=ourblue!80!black]
  (-4.4, 4.2) -- node[left=2pt, font=\scriptsize\itshape, text=ourblue, align=right]
  {info\\narrows} (-4.4, 0.4);
\node[sidebar, anchor=west] (cb1) at (4.6, 3.4)
  {graph view\\\textit{no} verbatim text};
\node[sidebar, anchor=west] (cb2) at (4.6, 1.9)
  {evidence index only\\\textit{\S7 ref [[64]] dropped}};
\node[sidebar, anchor=west] (cb3) at (4.6, 0.4)
  {bound quotes only\\\textit{\S2: 112 records $\to$ 87}};
\draw[gray!70, dashed, thin] (L1.east) -- (cb1.west);
\draw[gray!70, dashed, thin] (L2.east) -- (cb2.west);
\draw[gray!70, dashed, thin] (L3.east) -- (cb3.west);
\node[draw=ourblue, thick, rounded corners=2pt, fill=fsbg,
      align=center, inner sep=3pt, font=\scriptsize, below=3mm of L3]
  {RACE Overall $0.578$ ~|~ Insight $0.620$ \quad
   ($G_T$: 6 nodes, 594 records, 103 unique $[\![m]\!]$)};
\end{tikzpicture}
\caption{Three-layer writing pipeline on query~53. Information access narrows in lockstep with decision authority: Layer~1 sees the graph but no quotes; Layer~2 sees the section's evidence index but no quotes; Layer~3 sees the bound quotes only.}
\label{fig:case_q53_writing}
\end{figure*}

One paragraph from §2 reads (translated): ``\emph{NBIM's exclusion guidelines specify that companies with more than $30\%$ of revenue from thermal-coal extraction are subject to product-based exclusion}~$[\![19]\!]$\emph{; the same threshold applies to thermal-coal-fired power generation}~$[\![19]\!]\,[\![27]\!]$\emph{, with conduct-based exclusion further triggered by serious environmental damage or systematic human-rights violations}~$[\![21]\!]$.'' All three $[\![m]\!]$ markers resolve to NBIM's own \emph{Observation and Exclusion Guidelines} document and a Norwegian Ministry of Finance regulation, both retrieved on the \textsc{exploit}-routed follow-up of the previous subsection. The chain is therefore traceable end-to-end: Reader Part-A's verbatim quote $\to$ $\mathcal{E}$ insertion $\to$ SectionPlanner binding $\to$ SectionWriter rendering $\to$ post-filter audit.

\subsection{Why this case demonstrates all three loops}

The trajectory exhibits all three loops in a single cycle. \emph{Interpretive update} made the diagnosis possible: $e_2$'s \texttt{discovered\_items} list and the per-item co-occurrence pattern are recorded in the state by $\Gamma$, rather than re-derived on each turn; Reader Part-B \texttt{unexpected\_insights} (state pensions, shell-company GDP distortion) entered the planner's reflection alongside Part-A acceptance criteria. \emph{Deviation feedback} stopped the planner from grinding on $r_2$/$r_3$: the deviation profile carried both ``strong'' findings on the placeholder \emph{and} the absence of any per-entity distribution that downstream dimensions would have needed, exposing a structural rather than informational gap. \emph{Schema revision} converted the diagnosis into action: $R_\textup{conc}$ rewrote topology while preserving $f_2$ untouched, and the new topology compressed eight prospective tasks into four entity-focused investigations that all closed within one round each.

Deviation feedback manifests at both ends of this trajectory in opposite modes. At $R_\textup{conc}$ the typed deviation profile triggers a structural revision (Figure~\ref{fig:case_q53_planner_reflection}). At \texttt{finish()} the same profile distinguishes search-recoverable gaps from unrecoverable ones (structural absence, disclosure form, sensitivity, access barriers; Figure~\ref{fig:case_q53_termination}), stopping the search rather than continuing on \texttt{can\_improve=True} signals. The corresponding ablations on this query show $A_3$ failing the trigger entirely (no \texttt{discovered\_items} for the planner to read off) and $A_2$ running the original $4 \times 2 = 8$-task topology to budget exhaustion. This failure-mode contrast is reflected in the $-4.61$/$-4.87$~pp losses on the $p\geq2/3$ subset (\autoref{tab:ablation-accomm}).

\section{Use of Large Language Models}
\label{app:llm_use}

We use large language models in two distinct roles.

\emph{As components of the system under study}, LLMs serve as the planner, searcher, reader, PreFilter, writer, and cognitive graph manager agents within VeriTrace. Per-role backbone assignments and per-configuration details are reported in Table~\ref{tab:configs}, Appendix~\ref{app:prefilter}, and Appendix~\ref{app:reproducibility}.

\emph{As writing and presentation assistants}, the authors used LLMs in several auxiliary capacities: grammar polishing, table formatting, and sentence-level rephrasing in the manuscript; drafting and debugging code for the implementation; and assisting with figure generation, which the authors then reviewed and edited. All technical claims, methodological designs, derivations, and experimental analyses were authored and verified by human authors; no LLM-generated text or code was incorporated without verification against the underlying experiments, and no LLM was used to ideate the metacognitive framework or its theoretical motivation.

\section{Cost Analysis}
\label{app:cost}

We decompose end-to-end cost across VeriTrace's agent roles and writing
stages. We report per-query consumption and
US-dollar cost computed by applying Alibaba Cloud Bailian per-model
rates (Table~\ref{tab:cost_rates}) to the per-model token breakdown
released in \texttt{evidence/*.json}'s \texttt{usage\_stats}. Tokens
reflect prefill plus decode; input and output tokens are priced
separately. CNY rates are converted to USD at the May~2026 spot rate
of \textbf{7.2 CNY/USD}.

\begin{table}[h]
\centering
\scriptsize
\setlength{\tabcolsep}{4pt}
\renewcommand{\arraystretch}{1.0}
\begin{tabular}{lcccc}
\toprule
\textbf{Model} & \textbf{¥in/M} & \textbf{¥out/M} & \textbf{\$in/M} & \textbf{\$out/M} \\
\midrule
Q3.5-35B-A3B    & 0.4   & 3.2   & 0.056 & 0.444 \\
Q3.5-27B        & 0.6   & 4.8   & 0.083 & 0.667 \\
Q3.5-Plus       & 0.8   & 4.8   & 0.111 & 0.667 \\
Q3.5-122B-A10B  & 0.8   & 6.4   & 0.111 & 0.889 \\
Q3.6-Plus       & 2.0   & 12.0  & 0.278 & 1.667 \\
DSv4-Flash      & 1.0   & 2.0   & 0.139 & 0.278 \\
DSv4-Pro        & 12.0  & 24.0  & 1.667 & 3.333 \\
\bottomrule
\end{tabular}
\caption{Token prices per 1M tokens. Source: Alibaba Cloud Bailian
public pricing, regular (non-cached) rates as of May~2026. The USD columns
use 7.2~CNY/USD. Q = Qwen, DSv4 = DeepSeek-V4. Cached-input rates
(¥1/M for DSv4-Pro, ¥0.2/M for DSv4-Flash) are ignored here for
conservative reporting.}
\label{tab:cost_rates}
\end{table}

\begin{table*}[h]
\centering
\scriptsize
\setlength{\tabcolsep}{4pt}
\begin{tabular}{lcccccccc}
\toprule
\textbf{Config} & \textbf{LLM/q} & \textbf{in (K/q)} & \textbf{out (K/q)} & \textbf{Search/q} & \textbf{Wall/q (min)} & \textbf{\$in/q} & \textbf{\$out/q} & \textbf{Total \$/q (¥)} \\
\midrule
Config-3B           & 260 & 2261 & 1391 & 57.05 & 45.80 & 0.127 & 0.618 & 0.74 (¥5.35)  \\
Config-27B          & 212 & 1592 & 1106 & 47.31 & 57.41 & 0.112 & 0.521 & 0.63 (¥4.56)  \\
Config-Plus         & 264 & 2224 & 1348 & 59.45 & 64.78 & 0.284 & 1.102 & 1.39 (¥9.98)  \\
Config-DeepSeek     & 225 & 1750 & 461  & 46.59 & 40.61 & 1.461 & 0.495 & 1.96 (¥14.08) \\
\bottomrule
\end{tabular}
\caption{Per-query cost across configurations (means across 99--100
queries, calculated from their \texttt{usage\_stats}; rates from Table~\ref{tab:cost_rates}).
Config-DeepSeek uses DSv4-Pro for the Planner, Searcher, and Writer,
and DSv4-Flash for the Cognitive Graph Manager, Reader, and PreFilter.
Although DSv4-Pro is the most expensive component (15$\times$ Q3.6-Plus's
input rate, 2$\times$ output rate), the per-query cost stays moderate
because DSv4-Flash handles the high-volume Manager/Reader/PreFilter
roles.}
\label{tab:cost_usd}
\end{table*}

\paragraph{Cost composition for Config-DeepSeek.}
Of the \$1.96 per-query cost, the DSv4-Pro path (Planner, Searcher,
and Writer) contributes \$1.73 (88\%, \$1.33 input + \$0.40 output),
and the DSv4-Flash path (Cognitive Graph Manager, Reader, and PreFilter)
contributes \$0.23 (12\%). The input-dominated cost profile (75\%
input vs 25\% output) reflects the fact that Planner and Searcher
repeatedly re-prompt the same cognitive graph context across
multi-turn search dialogues. Enabling Bailian's cached-input rate
(¥1/M for DSv4-Pro, a 12$\times$ discount on cached prefixes) would
reduce overall input cost by approximately 50--80\% in practice
(depending on prefix reuse), lowering the realised per-query cost to
roughly \$0.7--\$1.2.

\section{Reproducibility Checklist}
\label{app:reproducibility}

\begin{strip}
\noindent Table~\ref{tab:repro} lists the artefacts and configuration parameters
released with the code repository; any exceptions are marked.

\medskip
\centering
\small
\setlength{\tabcolsep}{4pt}
\renewcommand{\arraystretch}{1}
\begin{tabular}{@{}p{0.29\linewidth} p{0.67\linewidth}@{}}
\toprule
\textbf{Component} & \textbf{Specification} \\
\midrule
\rowcolor{groupbg}\multicolumn{2}{l}{\emph{Models and APIs}} \\
Config-3B (all roles)            & Qwen3.5-35B-A3B (3B active / 35B total) \\
Config-27B                       & Planner/Searcher/CGM/Writer: Qwen3.5-27B (dense); Reader/PreFilter: Qwen3.5-35B-A3B \\
Config-Plus                      & Planner/Searcher/CGM: Qwen3.5-Plus (17B/397B); PreFilter: Qwen3.5-35B-A3B; Reader: Qwen3.5-122B-A10B (10B active); Writer: Qwen3.6-Plus (17B/397B) \\
Config-DeepSeek                  & Planner/Searcher/Writer: DeepSeek-V4-Pro (49B active / 1.6T total); Reader/PreFilter/CGM: DeepSeek-V4-Flash \\
LLM judge for RACE              & Gemini 2.5 Pro (\texttt{gemini-2.5-pro}, Vertex AI) \\
LLM judge for DeepConsult       & Gemini 2.5 Pro (\texttt{gemini-2.5-pro}, Vertex AI) \\
Search engine                   & Serper Google API (\texttt{SERPER\_API\_KEY}) \\
Page-content retrieval          & Jina Reader (\texttt{r.jina.ai}); PDF/Excel/CSV via dedicated parsers \\
\midrule
\rowcolor{groupbg}\multicolumn{2}{l}{\emph{Hyperparameters fixed across all runs}} \\
$\tau_h$ (Rel/Cred threshold)   & $3.5$ (between Likert ``adequate'' and ``good''); Appendix~\ref{app:strategies} \\
$\tau_l$ (low-quality threshold)& $2.5$; gates structural escalation \\
$\tau_\psi$ (unexpected threshold)& ``moderate or stronger'' on the $4$-level ordinal; gates the \textsc{Unex} predicate \\
$K$ (max attempts per edge)     & $3$; Definition~\ref{def:edge}, Appendix~\ref{app:update} \\
$T_\textup{max}$ (soft deadline)  & $70$ wall-clock minutes (\texttt{SOFT\_DEADLINE\_MINUTES}); independent iteration cap \texttt{max\_turn}$=20$ with $3\!\times$ hard ceiling; Appendix~\ref{app:termination} \\
Quality-profile cap             & $50$ most-recent records per node; Appendix~\ref{app:update} \\
Phase~2 orphan-repair rounds    & $\le 5$ before rollback; Appendix~\ref{app:safety} \\
SectionPlanner $|\mathcal{I}|$  & $[2, 5]$ ref-ids per insight \\
SectionWriter context window    & default $15$K-character tail (\texttt{PREVIOUS\_TEXT\_MAX\_CHARS}); per-language ceilings $20$K (CN) / $40$K (EN); Appendix~\ref{app:writing} \\
Reader verbatim quote length    & $200$--$500$ characters per finding \\
\midrule
\rowcolor{groupbg}\multicolumn{2}{l}{\emph{Datasets and protocols}} \\
DeepResearch Bench              & 100 queries (CN+EN), official RACE pipeline; \citet{du2025deepresearch} \\
DeepConsult                     & 102 queries; OpenAI DR reference reports as bundled in the DeepConsult release \citep{DeepConsult}; judge: Gemini 2.5 Pro \\
\bottomrule
\end{tabular}
\captionof{table}{Reproducibility checklist.}
\label{tab:repro}
\end{strip}

\section{DeepConsult Win/Tie/Loss Evaluation}
\label{app:Deepconsult}

\begin{table*}[h!]
\centering
\fontsize{8}{9}\selectfont
\setlength{\tabcolsep}{3pt}
\renewcommand{\arraystretch}{0.85}

\begin{tabularx}{\textwidth}{l l *{5}{>{\centering\arraybackslash}X}}
\toprule
\rowcolor{groupbg}
\textbf{Dim.} & \textbf{System} & \textbf{Win\%} & \textbf{Tie\%} & \textbf{Loss\%} & \textbf{Avg.} & \textbf{NWR} \\
\midrule

\multirow{4}{*}{\shortstack[l]{Instr.\\Follow.}}
 & Enterprise-DR & 53.9 & 20.6 & 25.5 & 4.89 & 0.679 \\
 & FS-Researcher & 69.6 &  8.8 & 21.6 & 5.37 & 0.763 \\
 & WebWeaver     & 67.6 & 17.6 & 14.7 & 5.72 & 0.821 \\
\rowcolor{ourbg}
 & \textbf{VeriTrace} & \textbf{73.5} & 15.7 & \textbf{10.8} & \textbf{6.00} & \textbf{0.872} \\
\midrule

\multirow{4}{*}{\shortstack[l]{Compre-\\hens.}}
 & Enterprise-DR & 71.6 & 13.7 & 14.7 & 6.56 & 0.829 \\
 & FS-Researcher & 77.5 &  3.9 & 18.6 & 6.78 & 0.806 \\
 & WebWeaver     & \textbf{90.2} &  5.9 &  3.9 & \textbf{7.76} & \textbf{0.958} \\
\rowcolor{ourbg}
 & \textbf{VeriTrace} & \textbf{90.2} & 5.9 & 3.9 & 7.31 & \textbf{0.958} \\
\midrule

\multirow{4}{*}{\shortstack[l]{Complete-\\ness}}
 & Enterprise-DR & 73.5 & 12.8 & 13.7 & 7.24 & 0.843 \\
 & FS-Researcher & 80.4 &  7.8 & 11.8 & 7.72 & 0.872 \\
 & WebWeaver     & 85.3 &  8.8 &  5.9 & 8.22 & 0.936 \\
\rowcolor{ourbg}
 & \textbf{VeriTrace} & \textbf{98.0} & 1.0 & \textbf{1.0} & \textbf{8.62} & \textbf{0.990} \\
\midrule

\multirow{4}{*}{\shortstack[l]{Writing\\Quality}}
 & Enterprise-DR & 59.8 & 17.6 & 22.6 & 5.76 & 0.726 \\
 & FS-Researcher & \textbf{63.7} & 14.7 & 21.6 & \textbf{6.10} & 0.747 \\
 & WebWeaver     & 57.8 & 19.6 & 22.6 & 5.56 & 0.720 \\
\rowcolor{ourbg}
 & \textbf{VeriTrace} & 62.8 & 13.7 & 23.5 & 5.85 & \textbf{0.727} \\
\midrule

\multirow{4}{*}{\textbf{Overall}}
 & Enterprise-DR & 64.7 & 16.2 & 19.1 & 6.11 & 0.769 \\
 & FS-Researcher & 72.8 &  8.8 & 18.4 & 6.49 & 0.797 \\
 & WebWeaver     & 75.2 & 13.0 & 11.8 & 6.81 & 0.859 \\
\rowcolor{ourbg}
 & \textbf{VeriTrace} & \textbf{81.1} & 9.1 & \textbf{9.8} & \textbf{6.94} & \textbf{0.887} \\
\bottomrule
\end{tabularx}

\caption{Per-dimension DeepConsult results under the matched Qwen3.5-27B backbone (102 queries; judge: \texttt{gemini-2.5-pro}). Win/Tie/Loss are percentages; Avg.\ is the 0--10 score, with 5 denoting a tie. Dimension-level NWR is computed as $\frac{\text{Win}}{\text{Win}+\text{Loss}}$, excluding ties; Overall NWR is the arithmetic mean of the four dimension-level NWRs rather than a pooled win rate. Bold indicates the best result within each dimension. VeriTrace achieves the best Overall NWR and leads on Instruction Following and Completeness; in Comprehensiveness, it ties WebWeaver on Win/Tie/Loss/NWR but has a lower Avg.\ by 0.45 points. Writing Quality is the only dimension where FS-Researcher leads on Win\% and Avg.}
\label{tab:deepconsult_perdim}
\end{table*}

We present per-dimension Win/Tie/Loss rates following the LLM-judged DeepConsult protocol \citep{DeepConsult} (Gemini 2.5 Pro, 102 queries), using the same evaluation pipeline as Table~\ref{tab:deepconsult_main}. Each row reports one system's pairwise results against the OpenAI Deep Research reference report. Baseline numbers are taken directly from their respective generated evaluation files, while the VeriTrace results are obtained from its corresponding evaluation bundle, as summarised in Table~\ref{tab:deepconsult_perdim}.

\section{Loop Comparison Across Deep Research Systems}
\label{app:closure_matrix}

The three loops introduced in \S\ref{sec:intro} (\emph{interpretive update}, \emph{deviation feedback}, and \emph{schema revision}) are not VeriTrace-specific design preferences but properties that, in our analysis, support a calibrated and evolving model of the task across long search horizons. Table~\ref{tab:closure_matrix} compares VeriTrace against the three open-source baselines along these three axes.

\begin{table*}[h!]
\centering
\small
\setlength{\tabcolsep}{4pt}
\renewcommand{\arraystretch}{1.2}
\begin{tabular}{@{}p{0.18\linewidth} p{0.25\linewidth} p{0.25\linewidth} p{0.25\linewidth}@{}}
\toprule
\textbf{System} & \textbf{Interpretive update} & \textbf{Deviation feedback} & \textbf{Schema revision} \\
\midrule
\rowcolor{ourbg}\textbf{VeriTrace (ours)}
  & \textbf{\ding{51}} ~Each finding is interpreted against the concept's current state: classified against the target concept's current criteria and accumulated evidence, not in isolation, and the result is written back to that concept's resolution state $s_j$ and residual criteria, so the system tracks how far each concept is resolved and re-targets search at the specific criteria it still leaves unsatisfied ($\Gamma$, \S\ref{sec:assimilation}, App.~\ref{app:update}).
  & \textbf{\ding{51}} ~Before each search the planner declares what the concept expects to obtain; the outcome is scored against that expectation into a typed deviation signal, which routes the planner to a matching correction strategy rather than an undifferentiated retry (\S\ref{sec:strategy}, App.~\ref{app:strategies}).
  & \textbf{\ding{51}} ~When accumulated feedback shows the framing itself diverges from reality, structural operators rewrite concepts and their relations while evidence is preserved (invariant \textbf{I1}), and the revised framing in turn reshapes the plan, so the agent escapes its initial framing ($R_\textup{conc}/R_\textup{pivot}/R_\textup{correct}$, App.~\ref{app:safety}).\\
\midrule
WebWeaver \citep{li2025webweaver}
  & \textbf{\ding{56}} ~Evidence updates the report outline without being interpreted against any concept's current state, and the outline carries no per-concept resolution state, let alone its unsatisfied criteria; the system tells only whether the outline looks complete, not how far each concept is resolved.
  & \textbf{\ding{56}} ~The planner only judges whether evidence is sufficient for a comprehensive outline; with no per-search declared expectation, outcomes are never scored into a typed deviation, and missing evidence triggers more search rather than a matching correction strategy.
  & \textbf{\ding{56}} ~The report outline is continuously restructured, but on report sections, not an explicit concept structure: structural revision is left to the LLM's implicit reasoning, with no operator that rewrites concepts and their inquiry relations while preserving evidence. Our $A_2$ ablation shows implicit reasoning alone is insufficient for such restructuring on harder queries (\S\ref{sec:ablation}). \\
\midrule
FS-Researcher \citep{zhu2026fsresearcher}
  & \textbf{\ding{56}} ~Evidence is filed into a topic-organised knowledge base without being interpreted against any concept's current state. Its only criteria are a static, task-level acceptance checklist (files present, format, coverage), not per-concept criteria that track what each concept still owes; so the system cannot tell how far any individual concept is resolved.
  & \textbf{\ding{56}} ~The Context Builder decides whether to keep searching from a coarse sense that information is still missing; it declares no per-search expectation and scores no outcome into a typed deviation, so a shortfall triggers another undifferentiated browsing round, not a matching correction strategy.
  & \textbf{\ding{56}} ~Todos and files can be added or reordered, but the concept structure of its knowledge is never restructured when the framing diverges from reality; revision stays at the task/file level, not the concept structure. \\
\midrule
Enterprise-DR \citep{prabhakar2025enterprise}
  & \textbf{\ding{56}} ~Reflection detects gaps and contradictions, but only over a global running summary with no explicit per-concept resolution state; these judgements stay at the summary level and never become an update to how far any individual concept is resolved, so the system stops when the summary looks complete rather than when each concept is actually resolved.
  & \textbf{\ding{56}} ~Reflection detects knowledge gaps in prose, but without a per-search declared expectation it cannot type the deviation or route a matching strategy; missing information triggers more search, not targeted correction.
  & \textbf{\ding{56}} ~Reflection re-plans a flat todo list, but the initial conceptual framing is never structurally rewritten when the framing diverges from reality; it cannot escape a wrong frame. \\
\bottomrule
\end{tabular}
\caption{Loop feature matrix: each deep research system against the three loops of VeriTrace. \textbf{\ding{51}}: realised as an explicit, typed mechanism; \ding{56}: absent or left to the LLM's implicit reasoning. The empirical consequences of each loop are isolated by ablations \textbf{A1} (deviation feedback), \textbf{A2} (schema revision), and \textbf{A3} (interpretive update) in \autoref{tab:ablation}.}
\label{tab:closure_matrix}
\end{table*}

\clearpage
\onecolumn
\raggedbottom
\section{Prompt Overview}
\label{app:prompts}
\FloatBarrier
 
\paragraph{Prompt summary.}
Table~\ref{tab:prompts_simple} summarises the twelve agent roles. Simplified versions of the prompts, including the \texttt{\{var\}} placeholders that the system substitutes at runtime, follow in the blocks below.
 
\begin{table}[!ht]
\centering
\scriptsize
\setlength{\tabcolsep}{4pt}
\renewcommand{\arraystretch}{1.05}
\begin{tabular}{p{0.27\linewidth} p{0.66\linewidth}}
\toprule
\textbf{Role} & \textbf{One-line job} \\
\midrule
\rowcolor{groupbg}\multicolumn{2}{l}{\emph{Search-side roles}} \\
1. Planner               & Reads the cognitive graph, picks one of \{\texttt{add\_task}, \texttt{propose\_restructure}, \texttt{finish}\}. \\
2. CGM (parse mode)      & Turns the user question into the initial concept-level cognitive graph $G_0$. \\
3. CGM (update mode)     & Implements $\Gamma$ (App.~\ref{app:update}): classifies each finding as \emph{criterion-satisfying}, \emph{redundant}, \emph{contradictory}, or \emph{unexpected}. \\
4. Searcher              & Runs \texttt{web\_search} / \texttt{multi\_search} / \texttt{select\_pages}; emits \texttt{[[m]]}-cited synthesis. \\
5. PreFilter             & Drops only off-topic or duplicate results; permissive by design. \\
6. Reader                & Extracts criterion-tagged findings with mandatory verbatim quotes (Part~A) plus unexpected insights (Part~B). \\
\midrule
\rowcolor{groupbg}\multicolumn{2}{l}{\emph{Verification + writing-side roles}} \\
7. Evidence Verifier     & Gate before $\mathcal{E}$: rejects or corrects answers not supported by the verbatim quote. \\
8. OutlinePlanner        & Layer~1: decides $S$ sections and the node subset $V_k \subseteq N_T$ each section covers. \\
9. SectionPlanner        & Layer~2: emits $(\textit{claim}, \mathcal{I})$ insight pairs with $|\mathcal{I}| \in [2,5]$ from $\mathcal{E}_k$. \\
10. SectionWriter        & Layer~3: writes one section at a time; only sees $\mathcal{E}_k$; \texttt{[[m]]} markers are post-filtered against bound evidence. \\
\midrule
\rowcolor{groupbg}\multicolumn{2}{l}{\emph{Restructuring (App.~\ref{app:safety})}} \\
11. Phase~1 (surgical)   & Applies $R_k \in \mathcal{A}_\textup{struct}$ on the modifiable subgraph; refuses under a multi-criterion rubric (e.g., $N_\textup{user}$ deletion, weak rationale, aggregation-only nodes). \\
12. Phase~2 (orphan repair) & Reattaches orphans via BFS up to five rounds; rolls back to $\widetilde{G}_t$ otherwise. \\
\bottomrule
\end{tabular}
\caption{One-line summary of the twelve VeriTrace agent prompts. Simplified prompt text is shown in the blocks below; placeholders \texttt{\{var\}} are substituted at runtime.}
\label{tab:prompts_simple}
\end{table}
\FloatBarrier

The blocks below reproduce a simplified version of each prompt: the complete control-flow logic is shown, while few-shot examples and verbose error-handling fragments (roughly 60\% of total prompt length) are omitted for space. The full prompts will be released with the code upon publication.

\begin{gmemorycaseboxfixed}{VeriTrace agent prompts --- Planner}
\begin{gmemcontent}
 
\noindent\textbf{1. Planner (CodeAct)} \textit{(\texttt{core/react\_prompt.py})}
\begin{redlog}
You are a search planner managing search tasks on a cognitive graph. Your goal
is to progressively build a knowledge graph to answer complex questions.
 
## Core Principle
Your sole mission is to fully and thoroughly answer the user's original
question. When findings or unexpected_discoveries reveal important content the
user did not anticipate (background factors, new research directions, specific
source types), actively follow those discoveries and dig deeper. The cognitive
graph externalises your current understanding -- it shows what you know, what
you don't know, and what to search next.
 
## Role Definition
You are the Architect of the search process. You do not search the web
directly; you maintain a graph object whose nodes are search tasks and whose
edges are dependencies. You iteratively expand the graph until all
cognitive-graph edges are explored or removed.
 
## Capability boundary (Searcher can ONLY ...)
1. Search the public web; 2. Read publicly accessible pages and PDFs.
The Searcher CANNOT: download files, perform page interactions
(clicks/forms/downloads), extract large structured datasets, access paywalled
or login-gated content, or call arbitrary APIs.
 
## Information-flow boundary
The cognitive graph is YOUR private workspace. The Searcher automatically
receives the direct parent node's findings as background, but cannot see the
full graph. Cognitive-graph content is ONLY updated through search results.
 
## API (sandboxed)
add_task(node_name, node_content, task_objective, attention_notes,
         is_verification, cognitive_edge_id, cognitive_target_id,
         task_type, specified_source, core_criteria, supplementary_criteria)
                                  # assimilation: dispatch a search task
propose_restructure(op_type, reasoning)
                                  # accommodation: R_k in A_struct
finish()                          # termination: subject to convergence guard
 
## Strategy selection (consume the deviation signal delta_j)
exploit | verify | substitute | pivot | explore  -- selected by pi(delta).
\end{redlog}
 
\end{gmemcontent}
\end{gmemorycaseboxfixed}

\begin{gmemorycaseboxfixed}{VeriTrace agent prompts --- Cognitive Graph Manager}
\begin{gmemcontent}
 
\noindent\textbf{2. Cognitive Graph Manager --- parse mode} \textit{(\texttt{cognitive\_graph\_manager.py})}
\begin{redlog}
You are a cognitive-graph parsing expert. Parse the user's question into a
concept-level cognitive graph -- extract research dimensions and conceptual
relationships, rather than predicting specific entities.
 
Output schema (JSON):
{
  "thinking_summary": "...",
  "entities": [{"id":"e1", "name":"...", "type":"start|placeholder",
                "cognitive_state":"START|UNKNOWN", "type_constraint":"...",
                "condition_constraints":[...], "hop_distance":int,
                "discovery_dependency":"e_<src>"}],
  "relations": [{"id":"r1", "source":"e_i", "target":"e_j", "type":"...",
                 "inquiry_goal":"a searchable question",
                 "state":"to_solve", "task_type":"open|specified",
                 "specified_source":null|"..."}],
   "structure": {"type":"serial|parallel|convergence|mixed",
                "parallel_chains":[...], "convergence_nodes":[...]}
}
\end{redlog}
 
\medskip
\noindent\textbf{3. Cognitive Graph Manager --- update mode (\S\ref{sec:assimilation}, $\Gamma$)} \textit{(\texttt{cognitive\_graph\_manager.py})}
 
\begin{redlog}
You are a Cognitive Graph Update Expert. Given the Searcher's response and
the current graph G_t, perform a four-step update on the target node n_j:
 
1. Extract discovered_items: organisations, products, or systems that
   directly answer the inquiry goal (exclude metadata).
2. Route findings by attribution: per-item attributes -> item_findings;
   patterns/rankings spanning items -> cross_item_findings.
3. Reconcile against acceptance criteria: residual unsatisfied criteria
   are written back as core_pending / supplementary_pending.
4. Branch divergent material:
   - Contradictions on the same criterion -> contradictions list as
     records {criterion, old_claim, new_claim, resolution, kept}.
   - Unexpected (relevant to original question AND outside criteria)
     -> unexpected_discoveries list.
   - Access failures or confirmed data absence -> search_experience
     (mutually exclusive with findings).
 
Cross-node routing: findings whose semantic content matches a non-target
node n_k are admitted to n_k via T_k^{t+1} = T_k^t U {task_t}; capped at
1-2 new edges per task. START nodes are immutable.
\end{redlog}
 
\end{gmemcontent}
\end{gmemorycaseboxfixed}

\begin{gmemorycaseboxfixed}{VeriTrace agent prompts --- Searcher and PreFilter}
\begin{gmemcontent}
 
\noindent\textbf{4. Searcher} \textit{(\texttt{graph.py})}
\begin{redlog}
You are a web search agent. Your task is to search for information and provide
comprehensive answers with citations.
 
## Available Tools
1. web_search:    {"query": "..."}                       # single query
2. multi_search:  {"queries": ["q1", "q2", "q3"]}        # max 5 queries; results
                                                          deduplicated and merged
3. select_pages:  {"indices": [0, 2, 5]}                 # dispatched to parallel
                                                          Reader agents
 
## Multi-search usage by task type
- SPECIFIED SOURCE: combine source expressions x topic keywords, e.g.
  ["site:williamreed.com Top 100 confectionery",
   "\"William Reed\" Top 100 confectionery companies",
   "\"William Reed Business Media\" confectionery ranking 2024"]
- OPEN EXPLORATION: use different angles -- keywords, perspectives, languages.
 
## Output
Return a synthesis that answers the search task with [[m]] citations indexed
to a local ref2url map; the orchestrator remaps these to global ref-ids.
\end{redlog}
 
\medskip
\noindent\textbf{5. PreFilter (Stage~1)} \textit{(\texttt{prefilter.py})}
\begin{redlog}
You are a search result filter. Your ONLY job is to remove results that are
completely off-topic or duplicate. When in doubt, KEEP the result -- the
Searcher needs to read broadly, including sources that might seem
low-authority, in order to verify the user's claims.
 
Inputs:  task content, acceptance criteria, list of (index, url, snippet).
Output:  {"read": [...indices...], "skip": [...indices...],
          "reasoning": "brief: why each skipped result is off-topic or dup"}
\end{redlog}

\end{gmemcontent}
\end{gmemorycaseboxfixed}

\begin{gmemorycaseboxfixed}{VeriTrace agent prompts --- Reader}
\begin{gmemcontent}

\noindent\textbf{6. Reader (dual-track Part~A + Part~B)} \textit{(\texttt{reader.py})}
\begin{redlog}
You are a precise information extractor. TWO tasks:
  1. Extract structured answers to acceptance criteria (Part A)
  2. Discover valuable information BEYOND the criteria  (Part B)
 
## Part A rules
- COMPLETENESS: extract ALL information the page contains; no shortcuts.
- DATA ACCURACY: when reading tables, list ALL column headers first, count
  from LEFT to RIGHT, include both row and column position when citing
  numbers.
- EVIDENCE REQUIREMENT (mandatory): every finding MUST include a direct,
  verbatim quote (200-500 characters) copied from the page. If you cannot
  find an exact quote that supports a finding, DO NOT report that finding.
  "evidence" is NOT your summary -- it is the page's own words.
- Two evidence types: descriptive (verbatim sentences) and table evidence
  (complete row + headers on next line).
 
## Part B rules
Surface unexpected_insights: facts that (i) help address the original
question's dimension and (ii) lie completely outside all acceptance
criteria. Look for: contradictions to Planner assumptions, important
unasked entities/relationships, timeline changes, follow-up leads.
 
## Output (flat JSON; not nested under part_a/part_b)
{"findings": {"<criterion>": {"answer":..., "evidence":...}},
 "gaps": [...],
 "unexpected_insights": [{"label":..., "insight":..., "evidence":...}],
 "quality_scores": {"currency":int, "relevance":int, ...},
 "temporal_context": "single string about page's effective time",
 "accessibility_notes": [{"criterion":..., "barrier":...,
                          "detail":..., "alternative_hint":...}]}
 
\end{redlog}
 
\end{gmemcontent}
\end{gmemorycaseboxfixed}
 
\begin{gmemorycaseboxfixed}{VeriTrace agent prompts --- Evidence Verifier and OutlinePlanner}
\begin{gmemcontent}
 
\noindent\textbf{7. Evidence Verifier} \textit{(\texttt{evidence\_verifier.py})}
\begin{redlog}
Check whether the answer matches the evidence, and correct it if there are
errors.
 
**Task goal**:           {task_goal}
**Acceptance criteria**: {criterion}
**Answer**:              {answer}
**Evidence**:            {evidence}
 
Output: a corrected answer if the original is unsupported or contradicted by
the evidence quotes; flag any hallucinations before findings are admitted to
the evidence store E.
\end{redlog}
 
\medskip
\noindent\textbf{8. OutlinePlanner (Layer~1)} \textit{(\texttt{outline\_planner.py})}
\begin{redlog}
You are a professional research report architect. Design a report outline
structure that directly answers the user's original question.
 
Inputs:
- Original question q
- Cognitive graph summary (only nodes with non-empty findings; empty roots
  are skipped)
- Evidence-availability index: per node, the distinct ref_idx values in E
 
Rules:
- One section per major aspect requested by the prompt.
- Bind each section to a node subset V_k of N_T (relevant_node_ids).
- Skip sections whose evidence is too thin to support analysis.
 
Output (JSON list of sections):
[{"section_id":int, "title":str, "description":str,
  "answers_aspect":str, "relevant_node_ids":[...]}]
\end{redlog}

\end{gmemcontent}
\end{gmemorycaseboxfixed}

\begin{gmemorycaseboxfixed}{VeriTrace agent prompts --- SectionPlanner}
\begin{gmemcontent}

\noindent\textbf{9. SectionPlanner (Layer~2)} \textit{(\texttt{section\_planner.py})}
\begin{redlog}
You are a senior research analyst. Plan insights and their evidence bindings
for ONE report section.
 
Inputs:
- The section spec (title, description, answers_aspect, V_k)
- Subgraph findings for V_k
- Evidence index for the section: {(ref_idx, criterion)} restricted to
  E_k = {m : task_m in T_k^T}
- previous_plans: insights of all earlier-numbered sections (no markers)
 
Rules:
- Each insight is a (claim, evidence_ids) pair.
- |evidence_ids| in [2, 5], drawn EXCLUSIVELY from E_k (no cross-section
  reuse).
- Skip any insight semantically subsumed by a prior one (cross-section
  deduplication).
\end{redlog}
 
\end{gmemcontent}
\end{gmemorycaseboxfixed}
 
\begin{gmemorycaseboxfixed}{VeriTrace agent prompts --- SectionWriter}
\begin{gmemcontent}
 
\noindent\textbf{10. SectionWriter (Layer~3)} \textit{(\texttt{section\_writer.py})}
\begin{redlog}
You are a senior research analyst and report writer. Write ONE high-quality
analytical section for the report.
 
Inputs you see:
- Original question
- Full report outline (so this section continues the narrative)
- Section spec + planned insights with bound evidence_ids
- Section evidence subset E_k (verbatim quotes v_m + finding summaries f_m)
- previous_text: 15K-char tail of earlier-written sections (continuity only)
 
Rules:
- Embed [[m]] citation markers for each cited evidence ref. Refs from other
  sections are NOT visible at generation time.
- A post-filter (FilterInvalidCitations) strips any [[m]] whose m is not in
  the section's bound evidence; removals are logged for audit.
- Write one claim per paragraph; do not repeat insights from previous_text.
\end{redlog}

\end{gmemcontent}
\end{gmemorycaseboxfixed}

\begin{gmemorycaseboxfixed}{VeriTrace agent prompts --- Restructuring}
\begin{gmemcontent}

\noindent\textbf{11. Restructuring Phase~1 (surgical)} \textit{(LLM stage of Appendix~\ref{app:safety})}
\begin{redlog}
You are a graph editor. Apply operator R_k in
{R_aug, R_prune, R_conc, R_pivot, R_correct} on the modifiable subgraph
(nodes outside N_user with empty findings, edges with status to_solve).
 
Evaluate each proposed change against a multi-criterion rubric.
Refusal conditions include (non-exhaustive):
- deletion target is in N_user (enforces I2);
- rationale cites local search difficulty ("search has been difficult",
  "no results found") rather than structural misalignment;
- merge would produce a pure aggregation node with no acceptance
  criteria of its own (Searcher cannot aggregate across arbitrary
  upstream nodes);
- proposed deletion contradicts evidence already in f_j^t (violates I1).
On refusal, the snapshot G_tilde is restored and the rejection reason
is returned to the Planner.
\end{redlog}
 
\medskip
\noindent\textbf{12. Restructuring Phase~2 (orphan repair)} \textit{(LLM stage of Appendix~\ref{app:safety})}
\begin{redlog}
After Phase~1, some nodes may be unreachable from the user-question root.
Detect orphans via BFS:  N_orphan = N(G') \ Reachable(root, E(G')).
 
Propose reattachment edges that connect each orphan to an existing reachable
node, preserving the protected sets P^N (nodes with findings or in N_user)
and P^E (edges with search history).
 
You have at most 5 rounds. If orphans remain after the 5th round, the entire
restructuring is rolled back to G_tilde.
\end{redlog}
 
\end{gmemcontent}
\end{gmemorycaseboxfixed}

\end{document}